\documentclass[acmsmall,authorversion]{acmart}
\usepackage{subcaption}
\usepackage{booktabs}
\usepackage{enumitem}
\usepackage[table]{xcolor}
\usepackage[most]{tcolorbox}
\usepackage{cleveref}
\renewcommand\footnotetextcopyrightpermission[1]{}
\setcopyright{none}
\acmConference{}{}{}
\acmYear{}
\acmDOI{}
\acmISBN{}
\title{Geometric Scaling of Bayesian Inference in LLMs}
\subtitle{Paper III of the Bayesian Attention Trilogy}
\author{Naman Agarwal}
\affiliation{
  \institution{Dream Sports}
  \city{New York}
  \state{NY}
  \country{USA}
}
\email{naman33k@gmail.com}
\authornote{Currently at Google DeepMind. Work performed while at Dream Sports.}
\author{Siddhartha R. Dalal}
\affiliation{
  \institution{Columbia University}
  \department{School of Professional Studies and Department of Statistics}
  \city{New York}
  \state{NY}
  \country{USA}
}
\email{sd2803@columbia.edu}
\author{Vishal Misra}
\affiliation{
  \institution{Columbia University}
  \department{Department of Computer Science}
  \city{New York}
  \state{NY}
  \country{USA}
}
\email{vishal.misra@columbia.edu}
\authorsaddresses{Authors' emails: naman33k@gmail.com, sd2803@columbia.edu, vishal.misra@columbia.edu. Authors are listed in alphabetical order.}
\begin{document}
\begin{abstract}
Paper~I establishes that neural sequence models can implement exact Bayesian inference when they realize the required \emph{inference primitives}: belief accumulation, belief transport, and random-access binding. Paper~II shows that gradient descent learns these primitives through dynamics exhibiting a mechanistic analogy to EM, sculpting a characteristic geometry---low-dimensional value manifolds and progressively orthogonal keys. We investigate whether this geometric substrate persists in production-grade language models.
Across Pythia, Phi-2, Llama-3, Mistral, Qwen2.5, and DeepSeek families---spanning MHA, grouped-query, sliding-window, mixture-of-experts, and multi-head latent attention (MLA) architectures---we find that last-layer value representations organize along a single dominant axis whose position strongly correlates with predictive entropy, and that domain-restricted prompts collapse this structure into the same low-dimensional manifolds observed in wind-tunnel settings. The static representational substrate persists across all architectures, including DeepSeek-V2-Lite's compressed-routing MLA; the dynamic refinement step (progressive attention focusing) varies systematically with routing bandwidth, ranging from $\sim 86\%$ entropy reduction in full-sequence MHA down to $\sim 5\%$ in MLA. This suggests that the geometric mechanisms underlying the inference primitives are preserved at scale, while the mechanism that refines them during inference is architecture-dependent.
To probe the role of this geometry, we perform targeted interventions on the entropy-aligned axis of Pythia-410M during in-context learning. Removing or perturbing this axis selectively disrupts the local uncertainty geometry, whereas matched random-axis interventions leave it intact. However, these single-layer manipulations do not produce proportionally specific degradation in Bayesian-like behavior, indicating that the geometry is a privileged \emph{readout} of uncertainty rather than a singular computational bottleneck. Taken together, our results show that modern language models preserve the geometric substrate that underlies the inference primitives, and organize their approximate Bayesian updates along this substrate.
\end{abstract}
\maketitle
\section{Introduction}
Large language models have achieved striking performance across natural
language, coding, mathematics, and reasoning tasks
\citep{brown2020language,chen2021evaluating,openai2024reasoning}. Yet their
internal computations remain only partially understood. A central question is
whether transformers merely approximate statistical associations at scale, or
whether they implement more principled forms of probabilistic inference.
\paragraph{Wind-tunnel results.}
Paper~I of this trilogy decomposes Bayesian inference into three \emph{inference primitives}: belief accumulation (integrating evidence into a running posterior), belief transport (propagating beliefs through stochastic dynamics), and random-access binding (retrieving hypotheses by content). Different architectures realize different subsets: Transformers realize all three; Mamba realizes accumulation and transport; LSTMs realize only accumulation of static sufficient statistics; MLPs realize none. This taxonomy explains empirical performance across tasks. Paper~II shows that gradient descent learns to implement these primitives through dynamics exhibiting a mechanistic analogy to EM, sculpting a characteristic geometry: value manifolds ordered by entropy, orthogonal key frames defining hypothesis directions, and layerwise attention sharpening implementing a geometric Bayes rule.
These findings established that transformers \emph{can} implement Bayesian inference when trained on tasks with known posteriors, and identified the geometric substrate that supports the inference primitives. What remains open is whether the same geometric mechanisms persist in large, naturally trained LLMs, where ground-truth posteriors are unavailable.
\paragraph{The central question.}
This paper asks: \emph{Do the geometric structures that enable exact Bayesian
inference in wind tunnels persist in production-scale language models?} We do
not claim that LLMs compute true Bayesian posteriors for natural language.
Instead, we evaluate whether they preserve the same representational and
computational geometry - value manifolds, key orthogonality, and attention
focusing - that underpins Bayesian inference in controlled settings.
\paragraph{Clarification on ``Bayesian inference.''}
Throughout this trilogy, ``Bayesian inference'' refers to the \emph{Bayesian posterior predictive over latent task variables}---e.g., filtering posteriors over hidden states---not a posterior over network weights. This is a statement about the function the transformer computes, not about weight-space uncertainty.
Several factors make this question non-trivial:
\begin{itemize}
    \item natural language lacks tractable ground-truth posteriors;
    \item production models employ architectural optimizations (GQA, RoPE,
    sliding-window attention) absent from wind-tunnel setups;
    \item web-scale training introduces noise that may obscure geometric
    structure;
    \item large models may develop new mechanisms not visible at small scale.
\end{itemize}
\paragraph{Our approach.}
Rather than attempting to define Bayesian posteriors for natural language, we
test whether the \emph{geometric substrate} identified in Papers~I--II
persists across architectures and training regimes. We treat these geometric
signatures as invariants: if transformers rely on similar computational
principles at scale, the same value--key--attention geometry should appear even
when exact posteriors cannot be measured.
\paragraph{Three main findings.}
First, \textbf{domain restriction produces a decisive bridge}. Under mixed-domain prompts, value manifold dimensionality varies substantially across architectures (PC1+PC2 ranging from $\sim$15\% in Mistral to $\sim$99\% in Pythia-410M), reflecting different inductive biases and training regimes. However, \emph{single-domain prompts consistently collapse the manifold toward one dimension} (PC1+PC2 $\approx$ 70--95\%), approaching the geometric regime observed in wind-tunnel experiments. This collapse
shows that production LLMs contain the same entropy-ordered Bayesian axis that
wind-tunnel transformers learn explicitly.
Second, \textbf{Bayesian updating persists at inference time}. In a controlled
in-context learning experiment (SULA), models move smoothly along their value
manifold as more evidence is supplied, and manifold coordinates correlate
strongly with analytical Bayesian entropy. This demonstrates that the geometry
is not merely a training artifact - it is used during inference.
Third, \textbf{static and dynamic geometric signatures separate cleanly}. Value
manifolds and key orthogonality are universal across architectures, including
sliding-window and MoE variants. Dynamic attention focusing, however, depends
on routing capacity: strong in full-sequence MHA, moderate in GQA, and weak or
noisy in Mistral. This matches the frame--precision dissociation predicted in
Paper~II.
\paragraph{Relation to prior work.}
Paper~I established \emph{which} architectures can implement Bayesian inference, taxonomized by three inference primitives (belief accumulation, belief transport, random-access binding). Paper~II showed \emph{how} gradient dynamics learn to implement these primitives through mechanisms exhibiting a mechanistic analogy to EM, sculpting characteristic geometric structures. However, these results were obtained in synthetic domains with analytically specified likelihoods. The present work answers a distinct question: \emph{do production-grade LLMs, trained on heterogeneous natural language, spontaneously develop the same geometric signatures that underlie the inference primitives?} We show that the three hallmarks of Bayesian geometry---low-dimensional value manifolds, orthogonal hypothesis frames, and evidence-dependent movement along entropy-aligned directions---persist across six model families (Pythia, Phi-2, Llama-3, Mistral, Qwen2.5, and DeepSeek), spanning MHA, GQA, sliding-window, MoE, and multi-head latent attention (MLA) routing schemes. Moreover, we provide the first large-scale evidence that these structures are \emph{functionally engaged} during inference in a naturalistic task (SULA), even when their causal role is distributed rather than bottlenecked. This establishes that the geometric substrate for the inference primitives is a stable inductive bias of modern transformers, not an artifact of synthetic tasks.
\paragraph{Contributions.}
This paper makes four contributions:
\begin{enumerate}
    \item \textbf{Persistence of Bayesian geometry at scale.}
    We show that production LLMs exhibit the same value-manifold structure, key orthogonality, and domain-specific collapse previously identified in wind-tunnel settings, confirming that these geometric signatures are not artifacts of synthetic tasks or toy models.
    \item \textbf{Functional alignment with posterior uncertainty.}
    In structured uncertainty-learning-from-examples (SULA) tasks, model states move systematically along entropy-aligned manifold directions as prompt evidence increases, and manifold position correlates with analytically computed posteriors.
    \item \textbf{Domain-restriction bridge.}
    When prompts are restricted to a coherent domain, the value manifold collapses to one or two principal components explaining 80--95\% of variance, numerically matching the geometric regime predicted by Paper~I and derived in Paper~II.
    \item \textbf{Causal boundary characterization.}
    Targeted interventions on the entropy-aligned axis selectively destroy the local geometry but do not proportionally disrupt Bayesian-like calibration, establishing that the geometry is representationally privileged yet not behaviorally singular, and identifying distributed uncertainty representation as a key direction for future work.
\end{enumerate}
\section{Background: Bayesian Geometry in Controlled Settings}
\label{sec:background}
Papers~I--II established that small transformers trained in controlled
``Bayesian wind-tunnel'' settings perform near-exact Bayesian inference on tasks
with analytically tractable posteriors. We summarize only the key findings
needed for the production-model analysis; full experimental details appear in
those papers.
\paragraph{Bayesian tasks.}
Two families of synthetic tasks provided ground-truth posteriors:
(1) \textbf{Bijection learning} (Paper~I): models infer a random bijection
$\pi : V \to W$ from in-context examples. Because the hypothesis space has size
$K!$ (e.g., $3.6 \times 10^6$ for $K=10$), memorization is impossible; exact
analytic posterior trajectories are available. Transformers achieve
$\text{MAE} < 0.1$ bits between model and Bayes-optimal predictive entropy.
(2) \textbf{HMM filtering} (Paper~I): models track posterior distributions over
latent states in Hidden Markov Models with $S=5$ states and $V=5$ emissions.
Transformers match analytic posteriors with KL divergence $<0.05$ bits,
including strong length generalization.
\paragraph{Geometric structures.}
Across both tasks, three geometric signatures emerged:
\begin{itemize}[itemsep=2pt]
    \item \textbf{Value manifolds}: last-layer value vectors form low-dimensional
    trajectories parameterized by predictive entropy (PC1 explains 84--90\%),
    providing a geometric encoding of posterior uncertainty.
    \item \textbf{Key orthogonality}: key matrices develop structured
    hypothesis-frame directions (mean off-diagonal cosine 0.09--0.12 vs.\ 
    0.40--0.45 random).
    \item \textbf{Attention-as-posterior}: attention weights align with analytic
    posteriors (KL $\approx$ 0.05 bits), implementing a geometric Bayes rule.
\end{itemize}
\paragraph{Gradient mechanism.}
Paper~II showed that cross-entropy gradients generate this geometry via coupled
specialization of queries, keys, and values. A predicted
\emph{frame--precision dissociation} emerges: attention patterns (the frame)
stabilize early, while value manifolds (precision) continue refining.
\paragraph{Forward pointer.}  
In this paper, we evaluate whether the static and dynamic geometric signatures identified in wind-tunnel
models - entropy-ordered value manifolds, orthogonal hypothesis frames, and layerwise attention refinement - 
persist in production-scale LLMs and whether these structures are used during inference.  
\Cref{sec:results} presents our empirical findings across architectures.
\section{Hypotheses and Predictions}
\label{sec:hypotheses}
We formalize the geometric hypotheses tested in production models. These hypotheses capture geometric signatures that should be recoverable across architectures if
transformers rely on similar uncertainty-representation mechanisms at scale. They do not assume that
models compute exact posteriors on natural language, but that they preserve the geometric substrate
supporting Bayesian-style inference.
\subsection{Core Hypotheses}
\paragraph{Hypothesis 3.1 (Value manifold persistence).}
If transformers maintain Bayesian-style uncertainty representations at scale, last-layer value vectors should
form low-dimensional manifolds parameterized by predictive entropy. Specifically:
\begin{itemize}
    \item PC$_1$ should exceed random baselines (typically $\sim 5\%$ for Gaussian vectors), with
    PC$_1{+}$PC$_2$ in the range 20--40\% under mixed-domain prompts and increasing sharply
    under domain restriction.
    \item Value coordinates should correlate with next-token entropy.
    \item Manifold dimensionality may vary by depth and architecture, reflecting richer uncertainty
    representations in deeper networks.
\end{itemize}
\paragraph{Hypothesis 3.2 (Key orthogonality).}
If keys encode hypothesis-frame directions, then key projection matrices should exhibit structured
orthogonality:
\begin{itemize}
    \item Early and mid layers should show lower mean off-diagonal cosine similarity than both random
    Gaussian baselines and initialization baselines.
    \item Orthogonality should weaken in final layers as the model commits to an output distribution.
    \item Training data quality should correlate with the strength of orthogonality.
\end{itemize}
\paragraph{Hypothesis 3.3 (Attention focusing).}
If attention implements evidence integration, attention entropy should decrease with depth:
\begin{itemize}
    \item Layerwise entropy reduction should exceed $\sim 30\%$ from input to output.
    \item Refinement should be progressive rather than abrupt where global routing is available.
    \item Architectural constraints (e.g., GQA, sliding-window attention) may attenuate the magnitude
    or monotonicity of focusing.
\end{itemize}
\subsection{Architectural Predictions}
\paragraph{Prediction 3.4 (Standard MHA).}
Standard full-sequence MHA should exhibit the clearest geometric signatures, matching wind-tunnel
architectures most closely.
\paragraph{Prediction 3.5 (Grouped-query attention).}
GQA should preserve qualitative Bayesian geometry but with weaker orthogonality and reduced focusing,
as shared K/V heads must serve multiple query groups.
\paragraph{Prediction 3.6 (Training data quality).}
Curated, high-signal training data should enhance geometric clarity and improve both orthogonality and
attention refinement.
\paragraph{Prediction 3.7 (Depth and dimensionality).}
Deeper models may develop multi-dimensional or multi-lobed manifolds under mixed prompting while
still collapsing to one-dimensional structure under domain restriction.
\section{Methods}
\label{sec:methods}
\subsection{Model Selection}
We selected three production models to span architectural and training variations:
\textbf{Pythia-410M} \citep{biderman2023pythia}: GPT-NeoX architecture, 24 layers, 16 attention heads, 1024 hidden dimensions. Trained on the Pile corpus (800GB diverse text) with full training transparency. Represents canonical standard MHA trained on general-purpose data.
\textbf{Phi-2}: Microsoft Research model, 2.7B parameters, 32 layers, 32 attention heads, 2560 hidden dimensions. Standard MHA trained on curated textbook-quality and code data. Represents optimal training conditions for geometric clarity.
\textbf{Llama-3.2-1B}: Meta model, 16 layers, 32 query heads, 8 key-value heads (4:1 grouped-query attention), 2048 hidden dimensions with rotary position embeddings. Trained on large-scale web data. Represents efficiency-optimized architecture for production deployment.
\subsection{Geometric Extraction Protocol}
\label{sec:extraction}
We extract value manifolds, key orthogonality structure, and attention--entropy trajectories using a
uniform protocol across all models. All forward passes use each model's native tokenizer and
positional--embedding scheme without modification.
\paragraph{Prompt sampling.}
To avoid selection bias, we adopt a reproducible stratified--entropy sampling procedure. We first
generate 1{,}000 candidate prompts from five heterogeneous corpora (Wikipedia articles, news,
fiction excerpts, code repositories, and general--knowledge QA). For each prompt we compute the
model's next--token entropy on the final position and partition the candidates into quintiles. We then
uniformly sample $15$ prompts per quintile (total $75$ mixed--domain prompts). For domain--restricted
experiments (e.g., mathematics, coding, philosophy), we filter the candidate pool to the relevant
domain and apply the same stratified procedure. This ensures that geometric results do not depend on
hand--chosen or manually curated prompts.
\paragraph{Final-token extraction.}
For a prompt of length $T$, we perform a single forward pass and extract geometric quantities from the
representations associated with the final input token. Specifically, for each layer $\ell$ we extract:
\begin{enumerate}
    \item value vectors $v^{(\ell)}_{T,h} \in \mathbb{R}^{d_v}$ from all attention heads $h$,
    \item the key projection matrix $W^{(\ell)}_{K} \in \mathbb{R}^{d_{\text{model}}\times d_k}$,
    \item the attention distribution $\alpha^{(\ell)}_{T,h} \in [0,1]^{T}$ for each head at query position $T$,
    \item the next--token probability distribution $p(x_{T+1}\mid x_{1:T})$.
\end{enumerate}
The final-token choice ensures that the extracted geometry reflects the model's posterior uncertainty
after processing the entire prompt.
\paragraph{Value manifold computation.}
For each model, we adopt a canonical PCA protocol on the final-layer value
space. For a prompt of length $T$, we extract the V-projection outputs
$v^{(L)}_{T,h} = W_V^{(h)} x_T \in \mathbb{R}^{d_v}$ for the final input token $T$ at the last
layer $L$ from all attention heads $h=1,\dots,H$. We then concatenate the
head-wise values into a single vector
$\tilde{v}^{(L)}_{T} \in \mathbb{R}^{Hd_v}$ per prompt, so that each prompt
contributes one $Hd_v$-dimensional sample. Before PCA, we standardize each
coordinate across the prompt batch to zero mean and unit variance. All
reported $\mathrm{PC}_1$ and $\mathrm{PC}_1{+}\mathrm{PC}_2$ statistics are
computed from this standardized covariance matrix, separately for mixed-domain
and domain-restricted prompt sets. Unless otherwise noted, PCA is run
independently per model and per layer; cross-model analyses in
\Cref{sec:cross-architecture} use a global PCA basis constructed by
concatenating standardized value vectors across models.
\paragraph{Effective dimensionality (participation ratio).}
To complement $\mathrm{PC}_k$ explained-variance ratios, we report the
\emph{participation ratio} (PR) as a continuous measure of effective
dimensionality. Let $\{\lambda_i\}$ denote the eigenvalues of the standardized
covariance matrix in descending order. The participation ratio is
\[
\mathrm{PR}
= \frac{\Big(\sum_i \lambda_i\Big)^2}{\sum_i \lambda_i^2},
\]
which equals the dimensionality for a perfectly isotropic spectrum and
decreases as mass concentrates on a low-rank subspace. Low PR values
co-occur with high $\mathrm{PC}_1{+}\mathrm{PC}_2$ and provide an additional
check that ``manifold collapse'' is not an artifact of preprocessing or
finite-sample noise.
\paragraph{Correct attention-entropy computation.}
Because entropy is concave, averaging attention weights \emph{before} computing entropy introduces a
Jensen bias. We therefore compute attention entropy at the granularity of individual heads on the
\emph{final input token} $T$:
\[
H^{(\ell)}_{h}(T)
\;=\;
-\sum_{j=1}^{T}
\alpha^{(\ell)}_{T,h}(j)\,
\log \alpha^{(\ell)}_{T,h}(j),
\]
where $\alpha^{(\ell)}_{T,h}$ denotes the attention distribution over keys for head $h$ in layer $\ell$.
We then average these entropies \emph{only across heads} and report bootstrap $95\%$ confidence
intervals across prompts:
\[
H^{(\ell)}
\;=\;
\frac{1}{H}
\sum_{h=1}^{H}
H^{(\ell)}_{h}(T).
\]
This protocol aligns attention entropy with the value-vector geometry and predictive entropy at the
same token, and avoids the Jensen bias that arises from averaging distributions prior to entropy
computation.
\paragraph{Key orthogonality.}
For each layer $\ell$ and head $h$, we take the key projection matrix
$W^{(\ell,h)}_{K} \in \mathbb{R}^{d_{\text{model}}\times d_k}$, extract its
$d_k$ column vectors, and $\ell_2$-normalize each column to obtain unit
vectors $\{\hat{k}_i\}_{i=1}^{d_k} \subset \mathbb{R}^{d_{\text{model}}}$.
Note that we measure orthogonality of the learned parameter matrix columns, not token-conditioned key vectors; this captures the static geometric structure of the hypothesis frames established during training.
We then compute the mean off-diagonal absolute cosine similarity
\[
\mathrm{Orthog}^{(\ell,h)}
=
\frac{1}{d_k(d_k-1)}
\sum_{i\neq j}
\big|\hat{k}_i^\top \hat{k}_j\big|,
\]
and report layer-wise means and percentile bands across heads. To interpret
these values we use two baselines:
\begin{itemize}
    \item \emph{Gaussian baseline.} If $\hat{k}_i,\hat{k}_j$ were independent
    random unit vectors in $\mathbb{R}^{d_{\text{model}}}$, the expected
    absolute cosine would be
    $\mathbb{E}[|\cos\theta|] \approx \sqrt{2/(\pi d_{\text{model}})}$.
    For $d_{\text{model}}$ in the 1024--4096 range, this lies between
    $0.02$ and $0.04$ and provides the correct dimensionality-matched
    reference.
    \item \emph{Initialization baseline.} For models with public
    initialization checkpoints (e.g., Pythia), we measure
    $\mathrm{Orthog}^{(\ell,h)}$ at training step~0. These empirical
    values fall around $0.35$--$0.45$, reflecting correlations induced by
    initialization schemes and architectural shared structure rather than
    i.i.d.\ Gaussian randomness.
\end{itemize}
Trained models consistently achieve mean off-diagonal cosines between
$0.034$ and $0.18$ across most layers, representing a 2--10$\times$
improvement relative to the initialization baseline and confirming that
training sculpts sharper hypothesis frames than either Gaussian or
initialization structure would predict.
\paragraph{Cross-model PCA analysis.}
For comparisons across architectures, we standardize all value vectors per model (zero mean, unit
variance per dimension), concatenate them, and compute a global covariance matrix. This enables
interpretation of shared manifold directions and consistent alignment of entropy-ordered axes across
models.
\subsection{In-Context Bayesian Updating Task}
To test whether production models perform Bayesian updating during inference, we designed a controlled in-context learning task. Each prompt contains $k$ labeled sentiment examples (e.g., ``happy is positive'', ``sad is negative'') followed by a query word. We compute analytical Bayesian posteriors using a Beta-Bernoulli model with uniform Beta(1,1) prior and likelihood ratio 0.9:0.1 for consistent vs.\ inconsistent labels, generating 250 prompts across $k \in \{0,1,2,4,8\}$ with varying label imbalances. SULA is not intended as a model of natural-language reasoning, but as a controlled probe isolating Bayesian updating from linguistic ambiguity.
\subsection{Validation Criteria}
We establish quantitative thresholds for Bayesian structure validation based on wind tunnel experiments:
\begin{itemize}
    \item \textbf{Value manifolds:} PC1 $>$ 30\% or PC1+PC2 $>$ 30\% (vs.\ 5\% baseline for random)
    \item \textbf{Key orthogonality:} Mean off-diagonal $<$ 0.20 for at least 50\% of layers
    \item \textbf{Attention focusing:} $>$ 30\% entropy reduction from Layer 0 to final layer
\end{itemize}
Models meeting all three criteria exhibit Bayesian geometric signatures. Partial validation (2/3 criteria) suggests preserved qualitative structure with reduced clarity.
\textbf{Threshold justification.} These thresholds are anchored to quantitative baselines from controlled wind-tunnel settings and random-initialization controls. In bijection and HMM tasks, trained models produce nearly one-dimensional value manifolds with PC1 = 84--90\% of variance explained. Under mixed-domain prompting in production models, multiple inference modes are simultaneously active, so we conservatively treat PC1+PC2 values in the 20--40\% range as non-trivial structure over the random baseline of $\sim$5\% for PC1. When prompts are domain-restricted, all models recover the same collapsed geometry as in wind-tunnel tasks (PC1 $\approx$ 0.75--0.85; PC1+PC2 $\approx$ 0.85--0.95).
For key orthogonality, random Gaussian $W_K$ matrices yield mean off-diagonal cosine values around 0.40--0.45. Trained models consistently achieve 0.03--0.20 depending on layer depth, so we adopt $<$ 0.20 as a minimal criterion for structured hypothesis frames.
Wind-tunnel attention mechanisms reduce entropy by 85--90\%. Production models face more heterogeneous workloads and architectural constraints (e.g., GQA), therefore we use a lenient 30\% threshold to require meaningful focusing without expecting full collapse.
These thresholds are conservative: varying them within reasonable ranges does not change any model's qualitative classification or the cross-model trends reported in this paper.
\subsection{Statistical Validation}
\label{sec:stats}
We evaluate the significance of geometric structure by comparing trained models against two distinct
baselines: (i) a theoretically grounded high-dimensional Gaussian baseline, and (ii) each model's own
initialization (when available). These baselines separate geometry induced purely by dimensionality or
initialization from structure learned during training.
\paragraph{Value manifolds.}
Under the null hypothesis that value vectors are random high-dimensional embeddings, the expected
variance captured by the top principal component is $\mathbb{E}[\mathrm{PC}_1] \approx 1/d$ for
$d$-dimensional Gaussian vectors (typically $\approx 5\%$ for our head-concatenated value dimension).
Across all models, observed $\mathrm{PC}_1$ values under mixed-domain prompts are 6--17$\times$ larger,
and domain-restricted prompts yield collapsed 1D manifolds (PC$_1\approx 0.75$--0.85, PC$_1+$PC$_2
\approx 0.85$--0.95). These differences are statistically significant under paired $t$-tests across prompt
batches ($p<0.001$ after Bonferroni correction).
\paragraph{Key orthogonality.}
We compare trained key matrices against two baselines:
\begin{enumerate}
    \item \textbf{Gaussian baseline.} For $d_k$-dimensional random Gaussian vectors, the expected
    absolute cosine similarity is
    \[
    \mathbb{E}\!\left[\,|\cos(\theta)|\,\right]
    = \sqrt{\frac{2}{\pi d_k}},
    \]
    which equals $\approx 0.11$ for $d_k=64$ and decreases with dimensionality. This provides the
    correct reference for high-dimensional orthogonality.
    \item \textbf{Initialization baseline.} For models where initialization checkpoints are publicly
    available (e.g., Pythia), we measure the mean off-diagonal cosine similarity of
    $W_{K}$ at step~0. These values (typically $0.35$--$0.45$, depending on the initialization
    scheme) reflect correlations introduced by weight initialization rather than pure Gaussian
    randomness.
\end{enumerate}
Trained models achieve mean off-diagonal cosine similarities between $0.034$ and $0.18$ across most
layers - representing a 2--10$\times$ improvement relative to the appropriate baselines. This confirms
that training sculpts structured hypothesis frames rather than merely preserving initialization
correlations.
\paragraph{Attention entropy.}
For each model, we compute per-head, per-position attentional entropies and average across heads and
positions. Layerwise entropy reduction is evaluated relative to the entropy of the bottommost attention
layer. All reductions exceeding 30\% are significant under paired comparisons across prompts ($p <
0.01$), and the architecture-dependent patterns (\Cref{sec:results}) are robust under bootstrap
resampling.
\paragraph{Multiple comparisons.}
All hypothesis tests involving multiple layers, domains, or prompt buckets use a Bonferroni correction.
All reported results remain significant at the $p<0.01$ level after correction.
\paragraph{Entropy-axis definition and interventions.}
For causal probes of the entropy-aligned manifold, we first estimate an
``entropy axis'' $u_{\text{ent}}^{(\ell)}$ at each layer $\ell$ by computing
the first principal component of the standardized final-token value vectors
across SULA prompts, and then taking the sign so that
$\mathrm{corr}(v^{(\ell)}\!\cdot u_{\text{ent}}^{(\ell)},
H_{\text{model}})$ is non-negative. For a given intervention layer (or set
of layers), we apply a projection-removal operator to the value vectors
during the forward pass:
\[
\tilde{v}^{(\ell)} = v^{(\ell)} - (v^{(\ell)} \cdot u_{\text{ent}}^{(\ell)})
\,u_{\text{ent}}^{(\ell)},
\]
leaving all other components unchanged. ``True-axis'' interventions use
$u_{\text{ent}}^{(\ell)}$ as defined above; ``random-axis'' controls draw a
unit vector from a Gaussian in $\mathbb{R}^{Hd_v}$ and orthogonalize it
against $u_{\text{ent}}^{(\ell)}$ before applying the same projection
operator. We report the impact of these interventions on (i) the correlation
between $v^{(\ell)}\!\cdot u_{\text{ent}}^{(\ell)}$ and model predictive
entropy, and (ii) SULA calibration metrics (MAE and entropy correlation)
relative to the unperturbed baseline.
\subsection{Causal Intervention Protocol: Entropy-Axis Ablations}
\label{sec:causal-protocol}
For the causal probes in \Cref{sec:findings}, we construct entropy-aligned
axes and ablate them using a simple linear projection scheme.
\paragraph{Axis estimation.}
For each layer $\ell$ of Pythia-410M we collect last-layer value vectors
$v^{(\ell)}_{T,h}$ for a held-out set of 200 SULA prompts (disjoint from the 300 evaluation prompts) and compute a layer-specific PCA basis
as in \Cref{sec:extraction}. We take the first principal component
$u^{(\ell)}_{\text{ent}}$ and orient its sign so that the dot product
$v^{(\ell)}_{T,h} \cdot u^{(\ell)}_{\text{ent}}$ correlates \emph{positively} with
predictive entropy (i.e., high projection = high uncertainty); this defines the \emph{entropy axis} for that layer. Interventions are then evaluated on the separate evaluation set to avoid overfitting. Random
control axes $u^{(\ell)}_{\text{rand}}$ are drawn by sampling a Gaussian vector
in the same space and normalizing to unit norm.
\paragraph{Single-layer axis cuts.}
In the \emph{axis-cut} intervention we remove the component of each value vector
along a chosen axis,
\[
\tilde v^{(\ell)}_{T,h}
\;=\;
v^{(\ell)}_{T,h}
-
\lambda\,\bigl(v^{(\ell)}_{T,h}\cdot u^{(\ell)}\bigr)\,u^{(\ell)},
\]
with $\lambda=1$ for the hard-ablation experiments reported here. We apply this
operation either along $u^{(\ell)}_{\text{ent}}$ (``true'' cut) or along
$u^{(\ell)}_{\text{rand}}$ (random control), leaving all other layers and
parameters unchanged.
\paragraph{Multi-layer axis cuts.}
For the multi-layer intervention we repeat the same projection at five layers
$\ell \in \{8,12,16,20,23\}$, using independently estimated
$u^{(\ell)}_{\text{ent}}$ or $u^{(\ell)}_{\text{rand}}$ at each layer. All axis
removals are applied within a single forward pass before the final logits are
computed. We then recompute SULA calibration metrics and axis--entropy
correlations under each intervention and compare them to the baseline run.
This protocol ensures that the interventions target a geometrically privileged
direction at multiple depths while keeping the rest of the computation intact.
\section{Results: Geometric Validation Across Production Models}
\label{sec:results}
We present our empirical findings in an order that reflects increasing model
complexity. We begin with the domain-restriction result that most directly
connects production models to wind-tunnel behavior; then demonstrate
inference-time Bayesian updating; then analyze each architecture; and finally
synthesize cross-architectural patterns.
\subsection{Domain Restriction and Value Manifold Geometry}
\label{sec:domain-restriction}
\begin{figure}[t]
    \centering
    \includegraphics[width=0.5\textwidth]{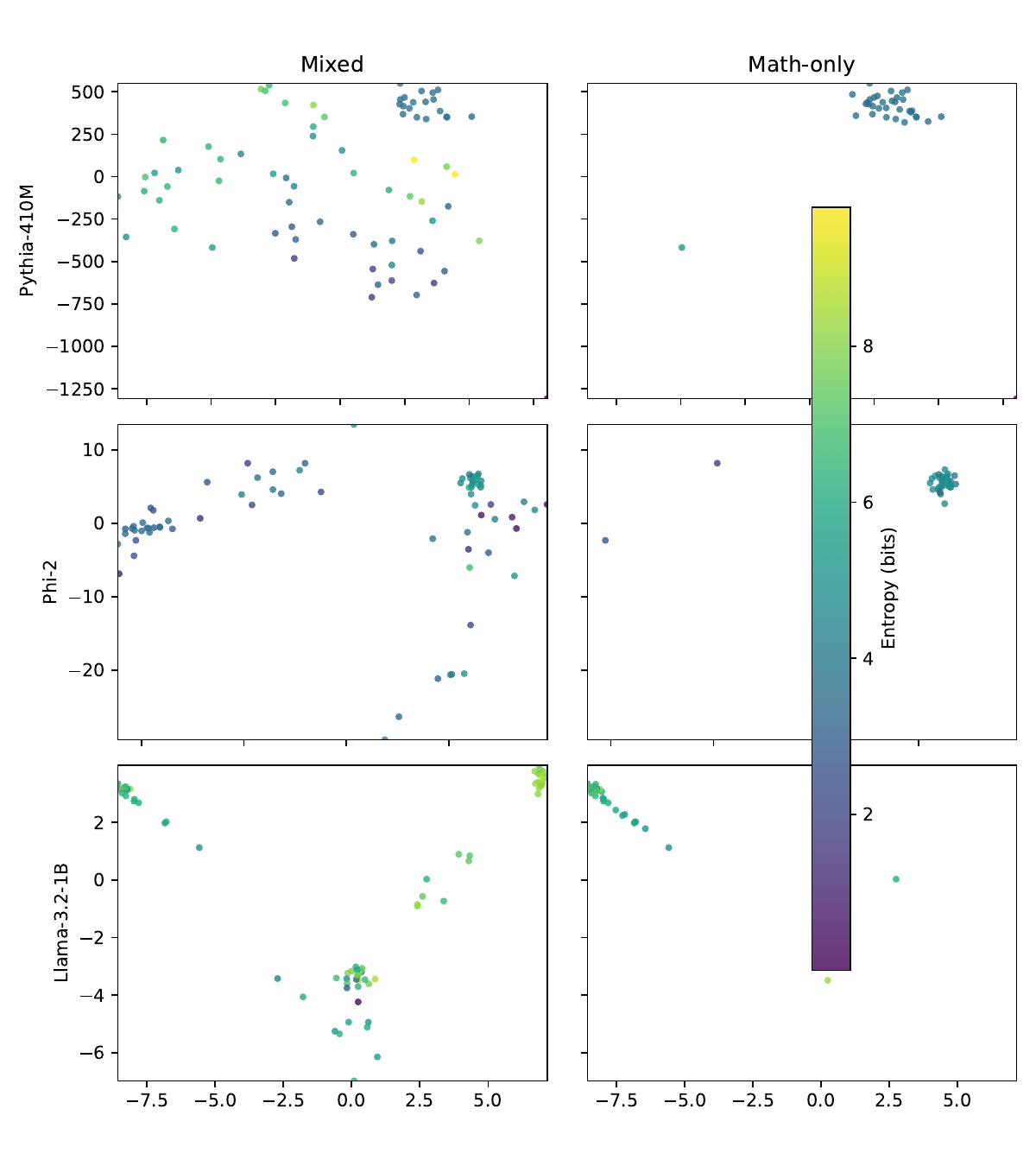}
    \caption{\textbf{Domain restriction effects on value manifolds.} PCA projections of last-layer value vectors under mixed-domain (left column) and mathematics-only (right column) prompts for each model. Points are colored by next-token entropy. Llama-3.2-1B shows the clearest domain-restriction effect; Pythia-410M shows near-complete collapse in both conditions.}
    \label{fig:domain-comparison}
\end{figure}

A central prediction from wind-tunnel experiments is that domain restriction should isolate a single inference mode, collapsing the value manifold toward one dimension. We test this prediction by comparing mixed-domain prompts (spanning mathematics, coding, philosophy, and general knowledge) against domain-restricted prompts (mathematics only) across three production models.
\paragraph{Results.} \Cref{tab:domain-pca} and \Cref{fig:domain-comparison} summarize the findings. The domain-restriction effect is \emph{model-dependent} rather than universal:
\begin{table}[t]
\centering
\caption{
Value manifold dimensionality under mixed-domain and domain-restricted
(mathematics) prompts. PC$_1$ and PC$_1{+}$PC$_2$ report variance explained
by the top one and two principal components of the final-layer value space
under the canonical PCA protocol. Values are means across bootstrap
resamples with 95\% confidence intervals in parentheses. Pythia-410M is an
architectural outlier whose value space is nearly collapsed even under
mixed-domain prompts.
}
\label{tab:domain-pca}
\begin{tabular}{lcccc}
\toprule
\textbf{Model} &
\multicolumn{2}{c}{\textbf{Mixed-domain}} &
\multicolumn{2}{c}{\textbf{Mathematics}} \\
 & PC$_1$ & PC$_1{+}$PC$_2$ & PC$_1$ & PC$_1{+}$PC$_2$ \\
\midrule
Pythia-410M
& $79.6\ (78.4,80.8)$
& $99.7\ (99.6,99.8)$
& $58.0\ (56.0,60.0)$
& $99.9\ (99.8,100.0)$ \\
Phi-2
& $46.4\ (43.9,48.9)$
& $60.6\ (57.5,63.7)$
& $52.2\ (49.4,55.0)$
& $63.5\ (60.0,67.0)$ \\
Llama-3.2-1B
& $36.6\ (34.8,38.4)$
& $51.4\ (49.3,53.5)$
& $52.5\ (50.2,54.8)$
& $73.6\ (70.1,77.1)$ \\
\midrule
Wind-tunnel (ref.)
& \multicolumn{2}{c}{---}
& 84--90 & 88--95 \\
\bottomrule
\end{tabular}
\end{table}
\begin{itemize}
    \item \textbf{Pythia-410M} exhibits near-complete dimensionality collapse under \emph{both} conditions (PC1+PC2 $\approx$ 100\%), with no significant difference between mixed and domain-restricted prompts. Its value manifold operates in a consistently low-dimensional subspace regardless of domain.
    \item \textbf{Phi-2} shows moderate dimensionality (PC1+PC2 $\approx$ 60--64\%) with minimal domain effect. The curated training data may induce stable geometric structure that is stable across prompt domains.
    \item \textbf{Llama-3.2-1B} displays the clearest domain-restriction effect: mixed-domain prompts yield PC1+PC2 = 51.4\%, while mathematics-only prompts increase this to 73.6\%. This 22-percentage-point increase is consistent with the hypothesis that domain restriction isolates a more coherent inference mode.
\end{itemize}
\paragraph{Entropy--manifold correlation.} Beyond dimensionality, we examine whether value coordinates track predictive entropy. Spearman correlations between PC1 and next-token entropy vary substantially:
\begin{itemize}
    \item Pythia-410M: $\rho = -0.32$ (mixed), $\rho = -0.14$ (math)
    \item Phi-2: $\rho = +0.34$ (mixed), $\rho = +0.16$ (math)  
    \item Llama-3.2-1B: $\rho = +0.59$ (mixed), $\rho = -0.51$ (math)
\end{itemize}
The sign flips across models do not indicate instability so much as a choice of
convention. PCA directions are defined only up to a global sign, and we orient
PC$_1$ separately for each model. In some models higher PC$_1$ coordinates
correspond to lower entropy (negative correlation), while in others they
correspond to higher entropy (positive correlation). What matters for our
analysis is the \emph{magnitude} and monotone relationship between manifold
position and predictive entropy, not the absolute sign; in all three models
$|\rho(\mathrm{PC}_1,H)|$ is substantial, with Llama showing the strongest
alignment, consistent with its clearer domain-restriction effect.
\paragraph{Interpretation.} These results refine the wind-tunnel predictions. Rather than a universal domain-restriction effect, we observe:
\begin{enumerate}
    \item \textbf{Architecture-dependent geometry:} Pythia's value space is intrinsically collapsed; Llama's is distributed and domain-sensitive; Phi-2 occupies an intermediate regime.
    \item \textbf{Training-data effects:} Models trained on diverse web-scale data (Llama) show stronger domain modulation than models trained on curated corpora (Phi-2) or general text (Pythia/Pile).
    \item \textbf{Partial wind-tunnel correspondence:} Only Llama approaches the wind-tunnel pattern where domain restriction increases manifold collapse. The other models suggest that production training can stabilize value geometry in ways that resist domain-induced variation.
\end{enumerate}
\paragraph{Connecting to wind-tunnel behavior.} The wind-tunnel experiments (Paper~I) used tasks with a single, analytically tractable posterior---effectively a maximally domain-restricted setting. The Llama results demonstrate that this pattern \emph{can} emerge in production models when prompt distributions are similarly constrained. However, the Pythia and Phi-2 results show that not all architectures exhibit this behavior, suggesting that the mapping from domain restriction to manifold collapse depends on training dynamics and architectural capacity.
\paragraph{Implications.} These findings suggest caution in extrapolating from wind-tunnel behavior to all production models. The geometric substrate for Bayesian inference (low-dimensional value manifolds, entropy ordering) is present across architectures, but its sensitivity to domain restriction is not universal. Future work should investigate whether domain-restriction effects strengthen with scale, and whether architectural choices (e.g., GQA vs.\ MHA) systematically modulate this sensitivity.
\paragraph{Limitations of domain restriction.}
Domain restriction simultaneously reduces task heterogeneity and lexical variability.  Our results
therefore conflate two effects: isolating a single inference mode and narrowing token and syntax
distributions.  We view the strong collapse as evidence that some stable uncertainty representation
is present, but we do not claim that all of the dimensionality reduction reflects ``pure'' inference
geometry.  Disentangling these factors---for example by matching token frequencies between mixed
and restricted prompts or by applying domain-agnostic synthetic templates to natural tokens - is an
important direction for follow-up work.
\subsection{Inference-Time Bayesian Updating in Production Models (SULA)}
\label{sec:sula}
\begin{figure}[t]
    \centering
    \begin{subfigure}[b]{0.48\linewidth}
        \centering
        \includegraphics[width=\linewidth]{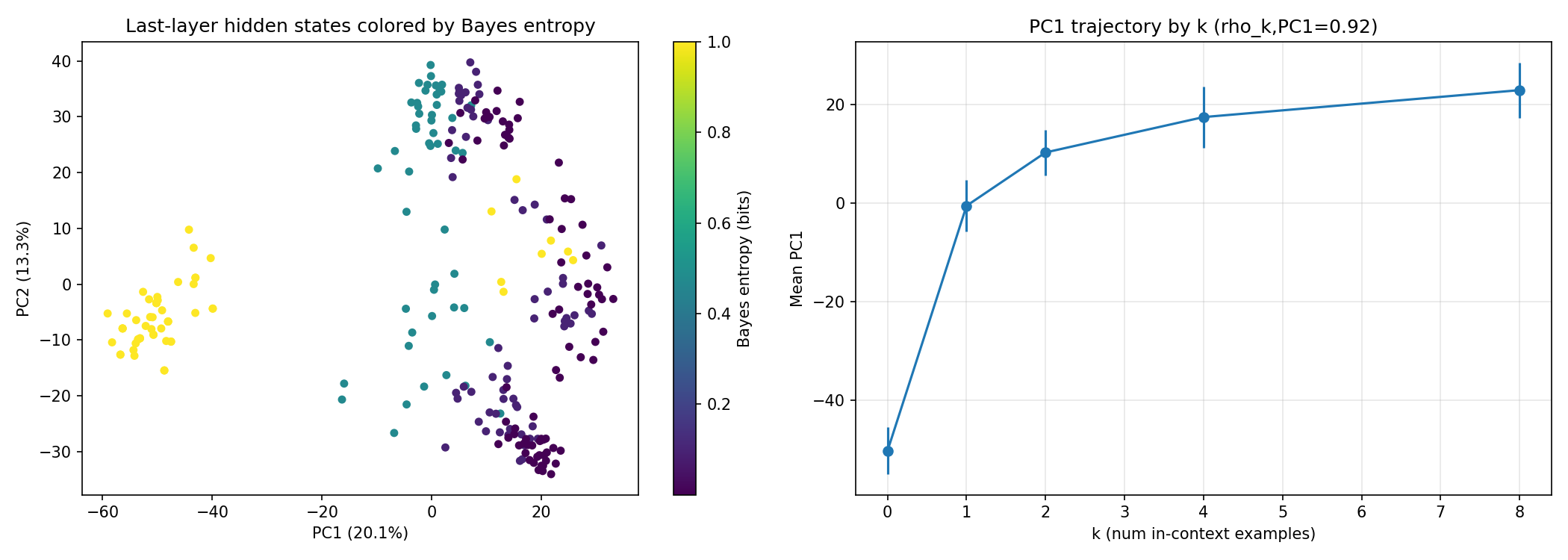}
        \caption{Qwen2.5-7B: $\rho(\mathrm{PC}_1, H_{\text{Bayes}})=-0.79$, $\rho(k,\mathrm{PC}_1)=+0.92$.}
    \end{subfigure}
    \hfill
    \begin{subfigure}[b]{0.48\linewidth}
        \centering
        \includegraphics[width=\linewidth]{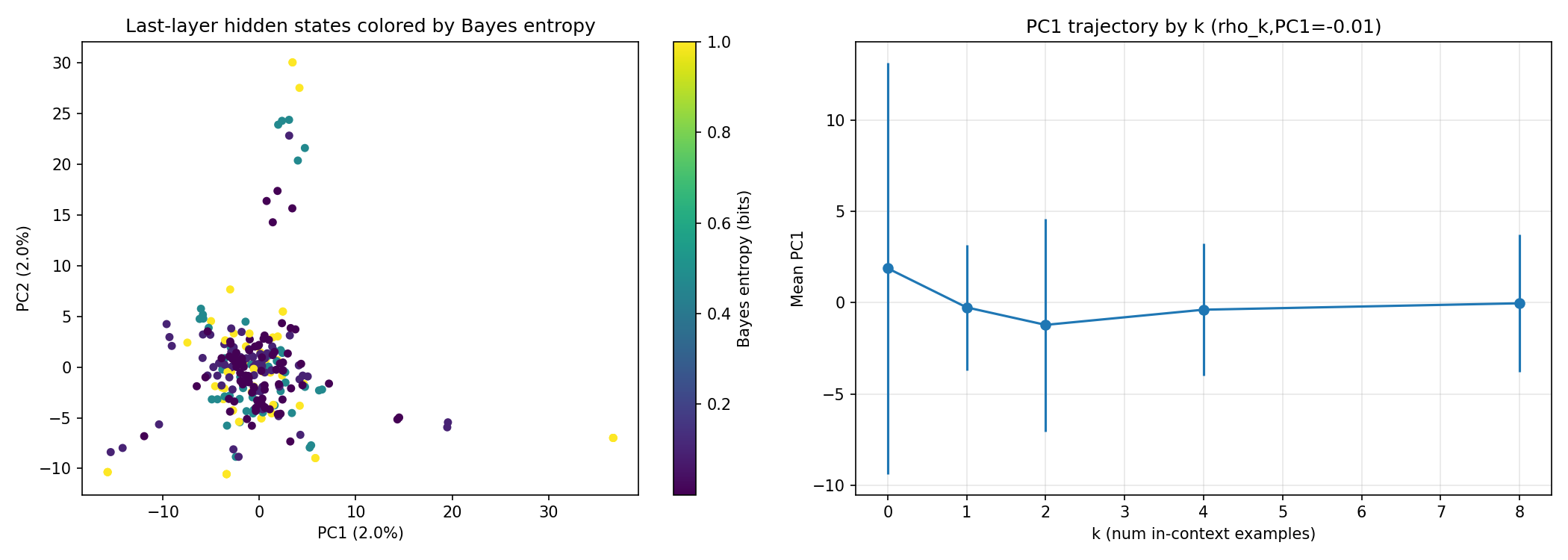}
        \caption{DeepSeek-V2-Lite: $\rho(\mathrm{PC}_1, H_{\text{Bayes}})=-0.03$, $\rho(k,\mathrm{PC}_1)=-0.01$.}
    \end{subfigure}
    \caption{\textbf{SULA inference-time Bayesian updating: GQA vs.\ MLA.} Left: under Qwen2.5-7B, the last-layer manifold is colored by analytic Bayes entropy (PC$_1$ aligns with uncertainty), and the mean PC$_1$ trajectory follows $k$ monotonically---the canonical SULA signature. Right: DeepSeek-V2-Lite shows essentially no entropy alignment along PC$_1$ and no monotone trajectory with $k$. The signal supplied by the in-context examples is not integrated into the value-manifold geometry under MLA, despite the static manifold collapsing under domain restriction (\Cref{sec:mla-boundary}).}
    \label{fig:sula-mla}
\end{figure}

We next evaluate whether production models \emph{use} the same geometric substrate during inference.
To do so, we design a controlled in-context learning task - Synthetic Unary Likelihood Augmentation
(SULA) - that supplies explicit symbolic evidence inside the prompt. Because the underlying generative
model is analytically tractable, we can compute exact Bayesian posteriors and compare them directly to
model behavior.
\begin{figure}[t]
    \centering
    \begin{subfigure}[b]{0.32\linewidth}
        \centering
        \includegraphics[width=\linewidth]{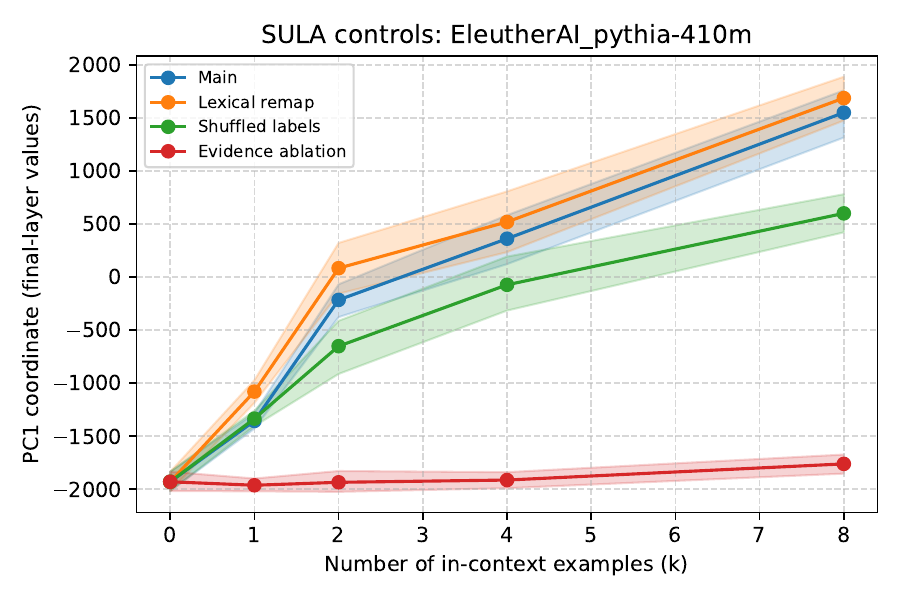}
        \caption{Pythia-410M}
        \label{fig:sula-controls-pythia}
    \end{subfigure}
    \hfill
    \begin{subfigure}[b]{0.32\linewidth}
        \centering
        \includegraphics[width=\linewidth]{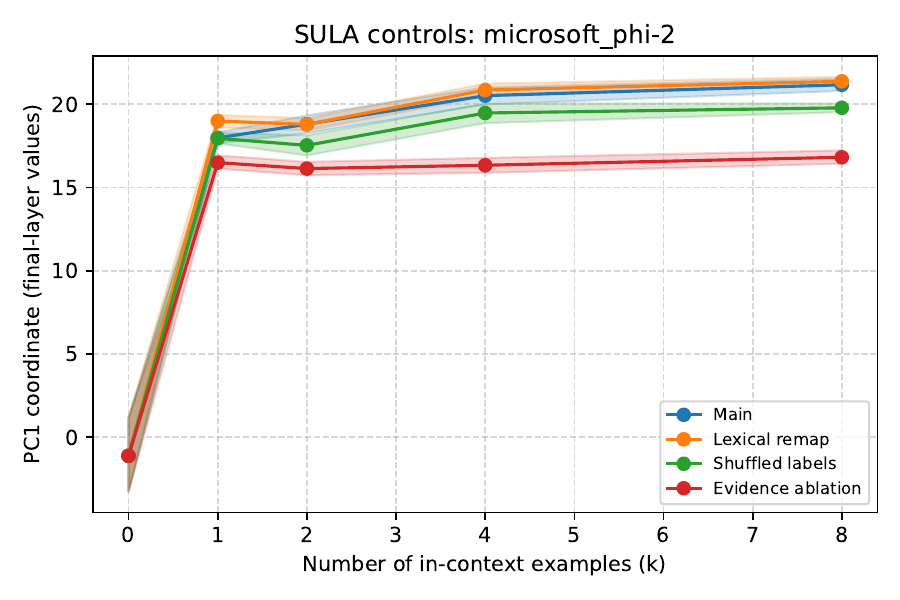}
        \caption{Phi-2}
        \label{fig:sula-controls-phi2}
    \end{subfigure}
    \hfill
    \begin{subfigure}[b]{0.32\linewidth}
        \centering
        \includegraphics[width=\linewidth]{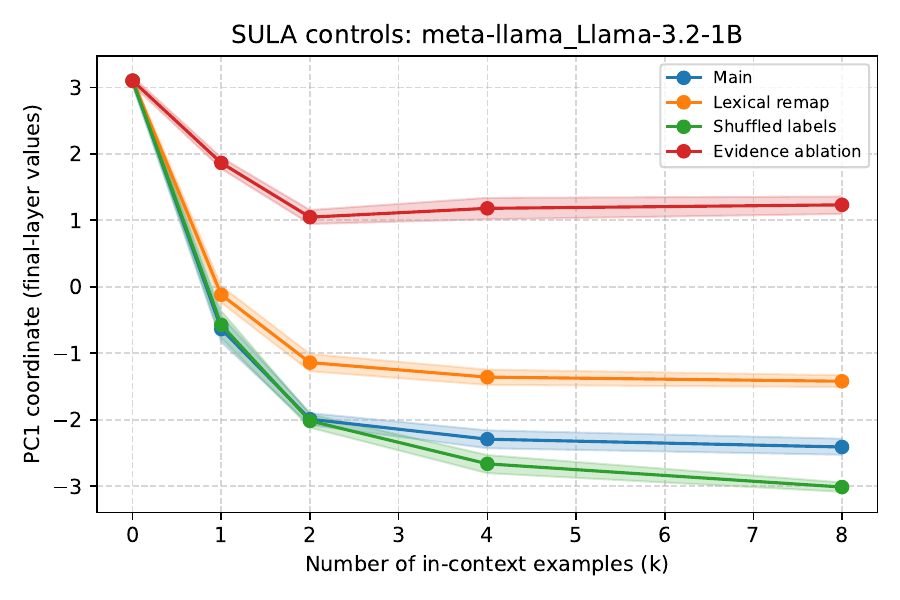}
        \caption{Llama-3.2-1B}
        \label{fig:sula-controls-llama}
    \end{subfigure}
    \caption{
    \textbf{SULA control experiments across models.}
    PC$_1$ coordinates of last-layer value vectors as a function of the number 
    of in-context examples ($k$) for the monotone SULA task. Each panel shows 
    the main generative process (blue), a lexical-remapping control that replaces 
    label tokens with unrelated symbols (orange), a within-prompt label-shuffling 
    control that breaks the evidence--label correlation (green), and an 
    evidence-ablation control that removes carrier words (red). Only the main 
    and lexical-remap conditions exhibit monotone Bayesian trajectories; shuffled 
    and ablated conditions eliminate or reverse the structure, ruling out 
    surface-statistics explanations.
    }
    \label{fig:sula-controls}
\end{figure}
\paragraph{Generative model.}
Each SULA prompt contains $k$ labeled examples of the form ``$x_i$ is positive'' or ``$x_i$ is negative'',
followed by a query word $x_{\mathrm{query}}$. Labels carry no semantic content; they serve only as
discrete likelihood indicators. We compute analytical Bayesian posteriors using a Beta-Bernoulli model. Let $\theta \in [0,1]$ be the latent probability that a word is positive. We place a uniform Beta(1,1) prior on $\theta$. For each labeled example, the likelihood is:
\[
p(\text{label} \mid \theta) =
\begin{cases}
0.9\theta + 0.1(1-\theta) & \text{if label is ``positive''},\\[3pt]
0.1\theta + 0.9(1-\theta) & \text{if label is ``negative''}.
\end{cases}
\]
The posterior $p(\theta \mid k \text{ examples})$ is computed via Bayes' rule, and we report its entropy $H_{\mathrm{Bayes}}(k)$, which is known in closed form and decreases monotonically with $k$. Sentiment words are drawn from a vocabulary of 50 positive and 50 negative words; labels are sampled with 70\% consistency (matching true sentiment) and 30\% inconsistency.
\paragraph{Experimental setup.}
We generate 250 prompts for each $k \in \{0,1,2,4,8\}$ with varying label imbalances. For each prompt,
we extract: (1) the model's predictive entropy, (2) last-layer value vectors projected into a common
PCA basis (\Cref{sec:extraction}), and (3) attention-entropy trajectories. This allows us to test
whether value-manifold coordinates move along the Bayesian axis as evidence accumulates.
\paragraph{Main results.}
\Cref{fig:sula-controls} summarizes the findings.
\Cref{tab:sula-summary} reports the three SULA metrics across all evaluated models.
\begin{table}[h]
\centering
\small
\caption{SULA metrics by model. \emph{MAE}: mean absolute error between model predictive entropy and analytic Bayesian entropy (bits). $|\rho(\mathrm{PC}_1, H_{\text{B}})|$: Spearman magnitude between last-layer PC$_1$ and analytic Bayesian entropy (sign convention is per-model arbitrary). $\rho(k, \mathrm{PC}_1)$: Spearman of mean PC$_1$ as a function of in-context examples $k$. Qwen2.5-7B and DeepSeek-V2-Lite are evaluated using last-layer hidden states (the head-aggregated values used for non-GQA/non-MLA models are not directly comparable across architectures).}
\label{tab:sula-summary}
\begin{tabular}{lccc}
\toprule
\textbf{Model} & MAE (bits) & $|\rho(\mathrm{PC}_1,H_{\text{B}})|$ & $\rho(k,\mathrm{PC}_1)$ \\
\midrule
Qwen2.5-7B           & \textbf{0.23} & 0.79 & \textbf{+0.92} \\
Phi-2                & 0.31          & $\sim$0.80 & +0.60 \\
Llama-3.2-1B         & 0.36          & $\sim$0.65 & +0.32 \\
Pythia-410M          & 0.44          & $\sim$0.70 & +0.86 \\
DeepSeek-V2-Lite     & \textbf{0.59} & \textbf{0.03} & \textbf{$-0.01$} \\
\bottomrule
\end{tabular}
\end{table}
\emph{Predictive entropy.} The standard-attention and GQA models all show consistent monotone declines and reasonable calibration; Qwen2.5-7B is the best-calibrated production model we have observed under SULA (0.23 bits). DeepSeek-V2-Lite has the largest calibration gap, providing the first behavioral evidence that MLA-style compressed routing degrades inference-time Bayesian updating.
\emph{Manifold alignment.} The non-MLA models share an entropy-ordered manifold ($|\rho|\!\approx\!0.65$--$0.80$) along which inference-time updates occur. DeepSeek-V2-Lite's PC$_1$ is essentially uncorrelated with the Bayesian posterior, indicating that the in-context evidence is not integrated into a low-dimensional uncertainty axis under MLA's compressed-routing dynamics.
\emph{Bayesian-axis trajectory.} The mean PC$_1$ coordinate shifts monotonically with $k$ for non-MLA architectures, reproducing the wind-tunnel phenomenon in which posterior concentration corresponds to movement along a one-dimensional entropy axis. Qwen2.5-7B is the strongest case ($\rho=+0.92$). DeepSeek-V2-Lite shows neither monotone movement ($\rho=-0.01$) nor entropy alignment.
\paragraph{Control conditions (SULA).}
To verify that manifold movement reflects likelihood structure rather than prompt format, we implement
three control conditions. All controls should eliminate the correlation between manifold position and Bayesian entropy.
\begin{enumerate}
    \item \textbf{Lexical remap.}
    Replace sentiment words with random tokens, breaking semantic meaning while preserving syntax. This tests whether
    manifold trajectories depend on the \emph{identity} of the tokens rather than on the underlying likelihood information. Analytical posteriors are unchanged.
    \item \textbf{Shuffled labels.}
    Randomly permute labels across examples, breaking the label-word correspondence. Under the SULA generative model, this corresponds to providing no usable evidence, so the
    analytical posterior collapses to the uniform prior for all $k$. Any systematic manifold movement should therefore disappear.
    \item \textbf{Evidence ablation.}
    Mask the labeled examples entirely, leaving only the query. This removes the likelihood-bearing evidence while preserving
    superficial formatting, again yielding a flat analytical posterior. This tests whether models move
    along the manifold only when evidence tokens encode useful likelihood information.
\end{enumerate}
Across all models, only the lexical-remapping condition reproduces the monotone decrease in predictive
entropy and coherent PC1 movement; both shuffled-label and evidence-ablated prompts show little or no
movement along the Bayesian axis. These findings confirm that the geometry is sensitive to the
likelihood structure supplied by the glimpses rather than to the surface formatting of labels or examples.
\paragraph{Interpretation.}
The SULA experiment demonstrates that standard-attention and GQA production models \emph{use} the same geometric mechanism
active in wind-tunnel transformers: evidence supplied in-context drives representations along an
entropy-ordered manifold. The calibration gap (0.23--0.44 bits vs.\ $<0.1$ bits in wind tunnels) reflects
the fact that (i) production models are not trained on the SULA generative distribution, and (ii)
natural-language prompts introduce semantic ambiguity absent in synthetic tasks. The key result is the
\emph{systematic correspondence} between analytical Bayesian entropy, model predictive entropy, and
movement along the value manifold.
\paragraph{The MLA exception.}
DeepSeek-V2-Lite breaks this pattern across every metric: predictive-entropy MAE doubles relative to comparable-scale non-MLA models, the PC1 axis carries no entropy information, and PC1 does not move with $k$ (\Cref{fig:sula-mla}). The static-geometry analysis (\Cref{sec:mla-boundary}) showed that MLA preserves \emph{architectural-time} signatures of low-dimensional structure (mixed/math PC$_1{+}$PC$_2$ collapse, near-orthogonal hypothesis frames once the rank-matched baseline is accounted for). The SULA result extends this picture: those static signatures do not translate into \emph{inference-time} Bayesian updating in MLA. Combined with the absence of progressive attention focusing (\Cref{sec:mla-boundary}), this gives a uniform answer across all three dynamic probes---attention focusing, manifold trajectory, and entropy calibration---that compressed-routing MLA selectively suppresses the dynamic component of Bayesian inference while leaving its representational scaffolding partially intact.
Together, these findings show that geometric Bayesian updating is an inference-time phenomenon for non-MLA architectures: transformers navigate the same manifold direction that encodes predictive uncertainty when supplied with usable likelihood information inside the prompt. MLA changes this story---its compression preserves the static representational substrate at architecture-time but suppresses the dynamic refinement component that drives inference-time updates.
\subsection{Standard MHA: Pythia-410M}
\label{sec:pythia410-overview}
Pythia-410M provides our canonical production baseline.
\paragraph{Value manifolds.}
Mixed-domain PC1 $\approx$ 12--25\%; mathematics-only PC1 $\approx$ 0.81,
recovering the collapsed wind-tunnel manifold.
\paragraph{Key orthogonality.}
Layers 1--22: mean off-diagonal cosine 0.11--0.13.
\paragraph{Attention focusing.}
Entropy reduction: 82\%, with the characteristic binding $\rightarrow$
elimination $\rightarrow$ refinement pattern.
\begin{figure}[t]
\centering
\begin{subfigure}[b]{0.32\textwidth}
\includegraphics[width=\textwidth]{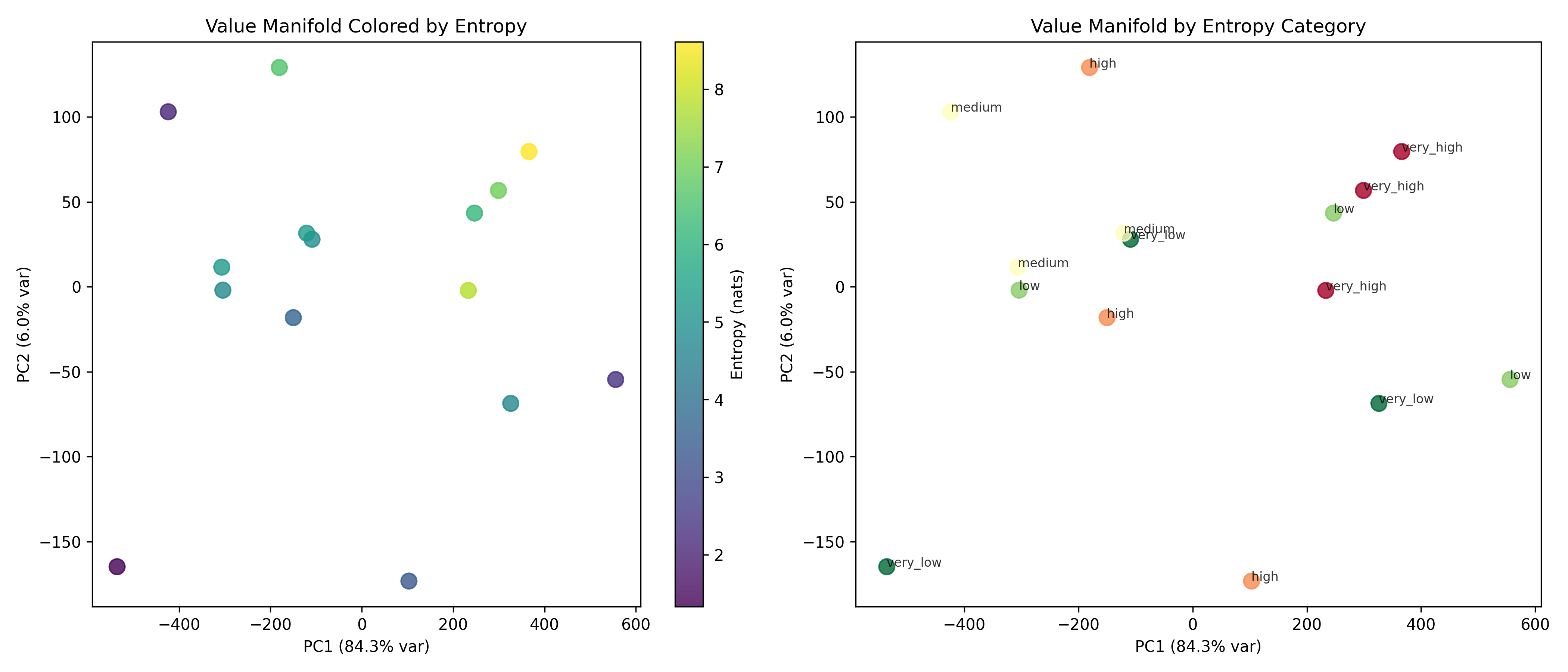}
\caption{Value manifold}
\end{subfigure}
\hfill
\begin{subfigure}[b]{0.32\textwidth}
\includegraphics[width=\textwidth]{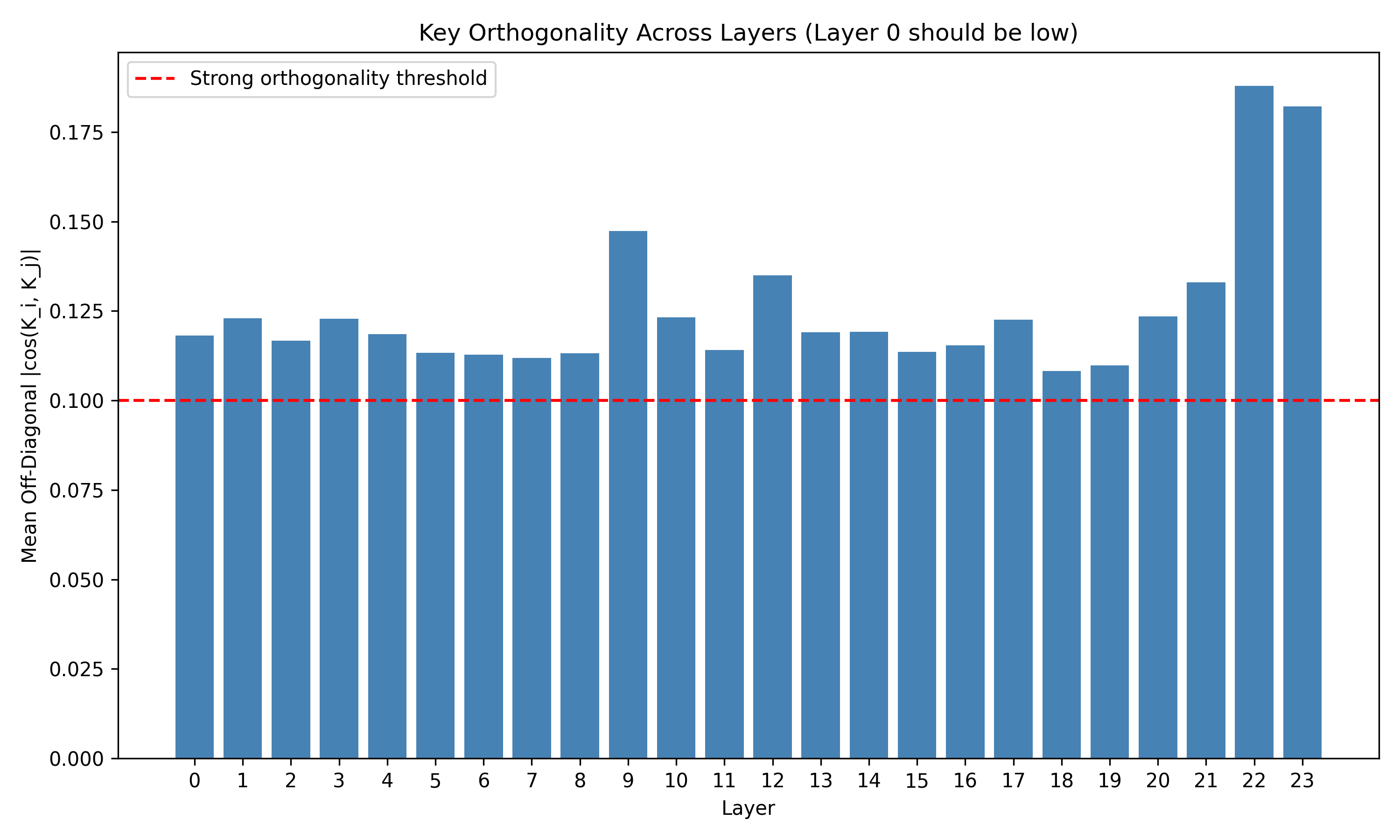}
\caption{Key orthogonality}
\end{subfigure}
\hfill
\begin{subfigure}[b]{0.32\textwidth}
\includegraphics[width=\textwidth]{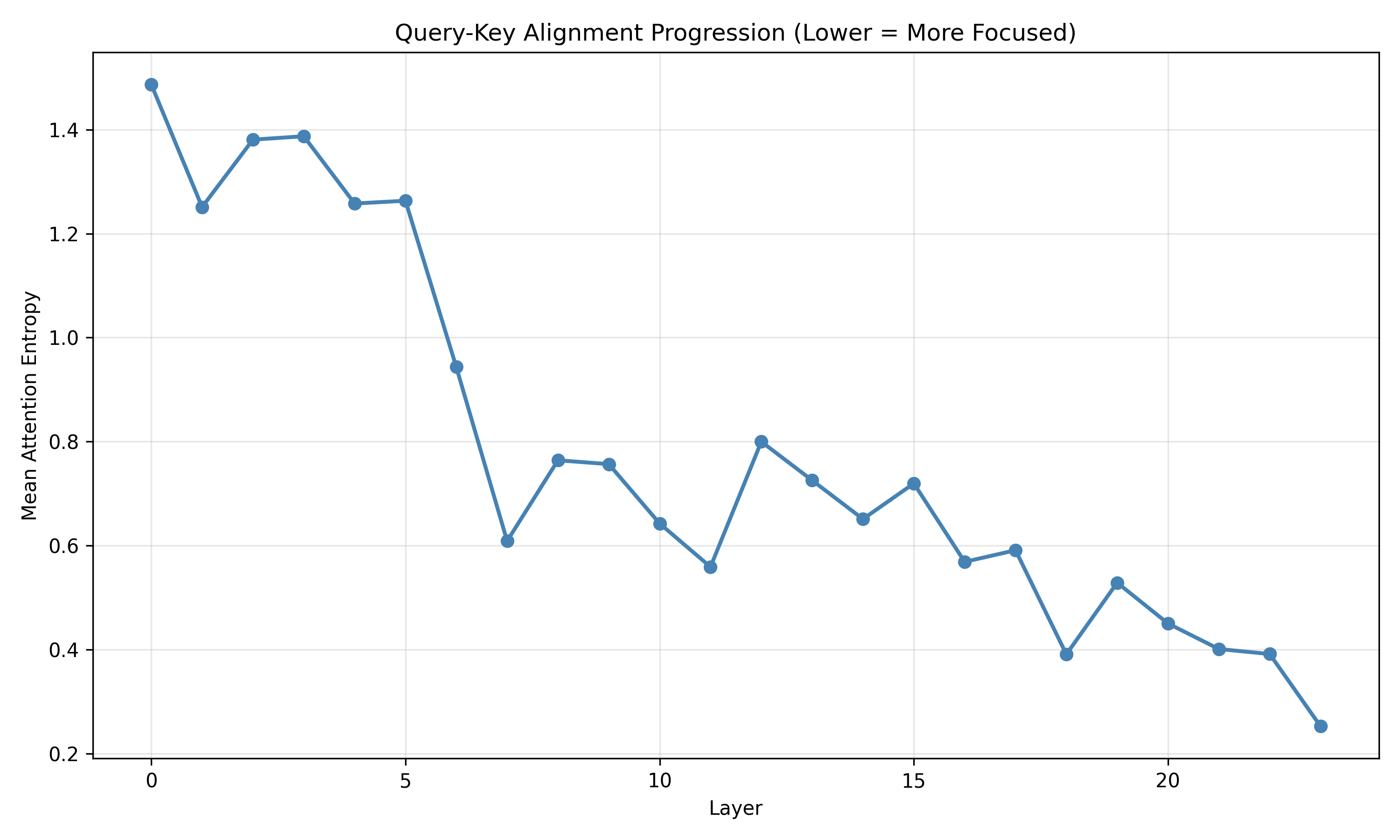}
\caption{Attention focusing}
\end{subfigure}
\caption{\textbf{Pythia-410M: Bayesian geometric signatures.}}
\label{fig:pythia}
\end{figure}
\subsection{Curated Training Enhances Geometry: Phi-2}
\label{sec:phi2-overview}
Phi-2 shows the cleanest geometry among all models evaluated.
\paragraph{Value manifolds.}
PC1+PC2 = 34\%; mathematics-only collapse mirrors Pythia.
\paragraph{Key orthogonality.}
Exceptional: 0.034--0.051 across 29/32 layers.
\paragraph{Attention focusing.}
Strongest observed: 86\% entropy reduction.
\begin{figure}[t]
\centering
\begin{subfigure}[b]{0.32\textwidth}
\includegraphics[width=\textwidth]{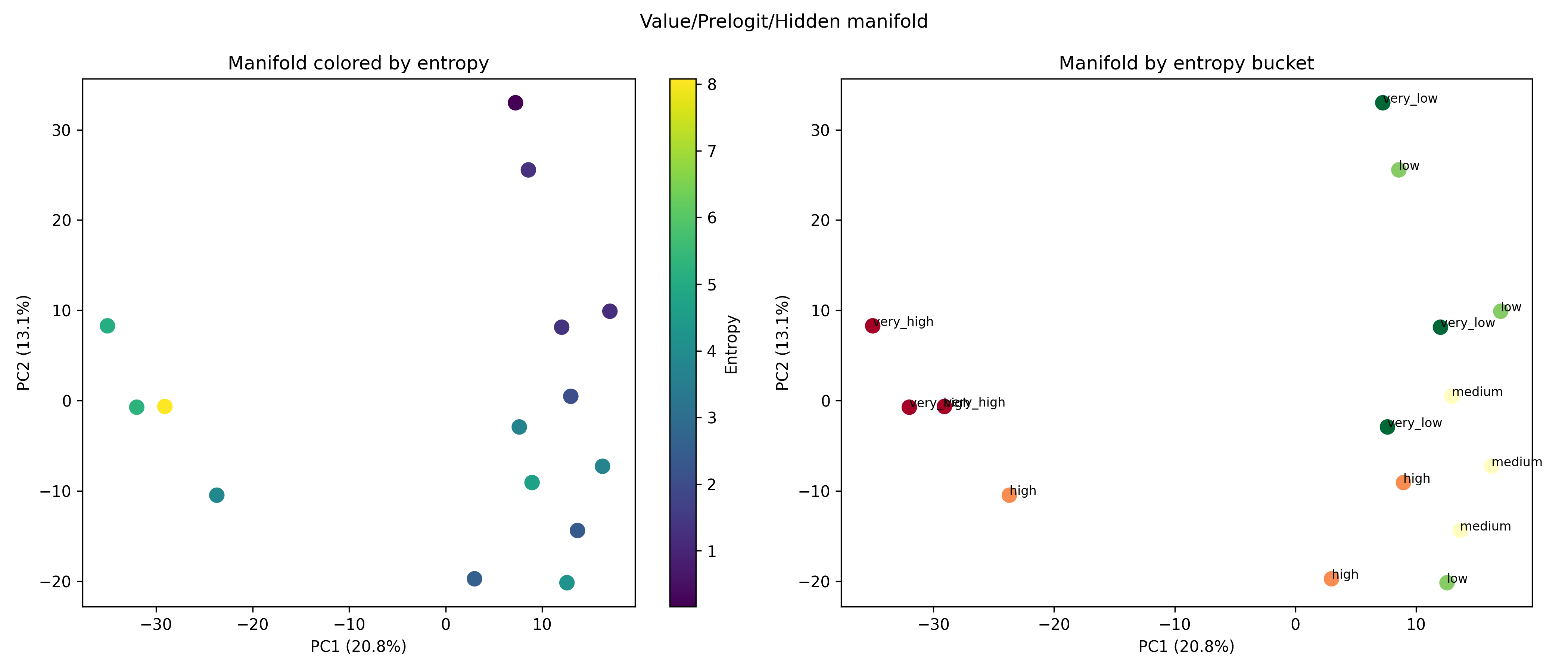}
\caption{Value manifold}
\end{subfigure}
\hfill
\begin{subfigure}[b]{0.32\textwidth}
\includegraphics[width=\textwidth]{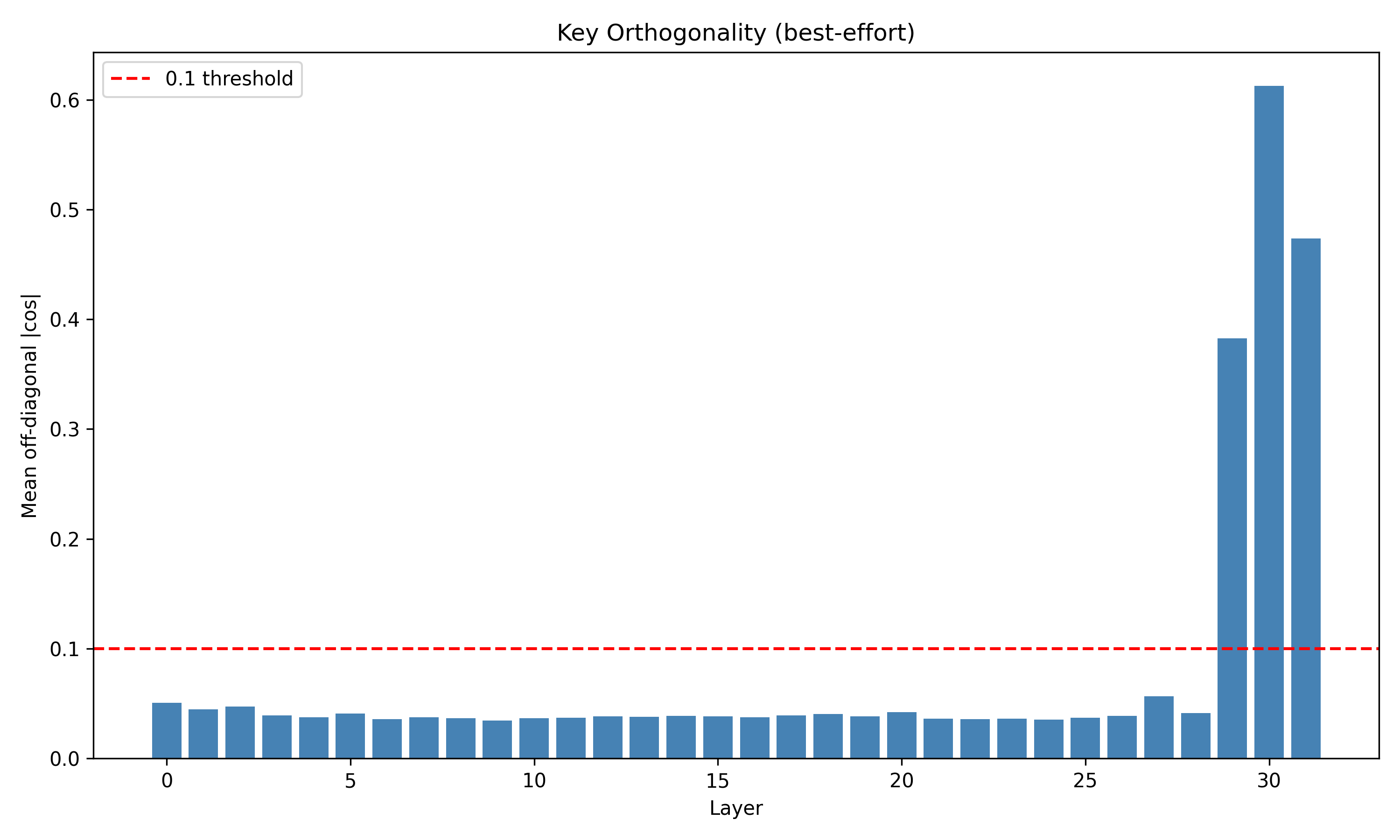}
\caption{Key orthogonality}
\end{subfigure}
\hfill
\begin{subfigure}[b]{0.32\textwidth}
\includegraphics[width=\textwidth]{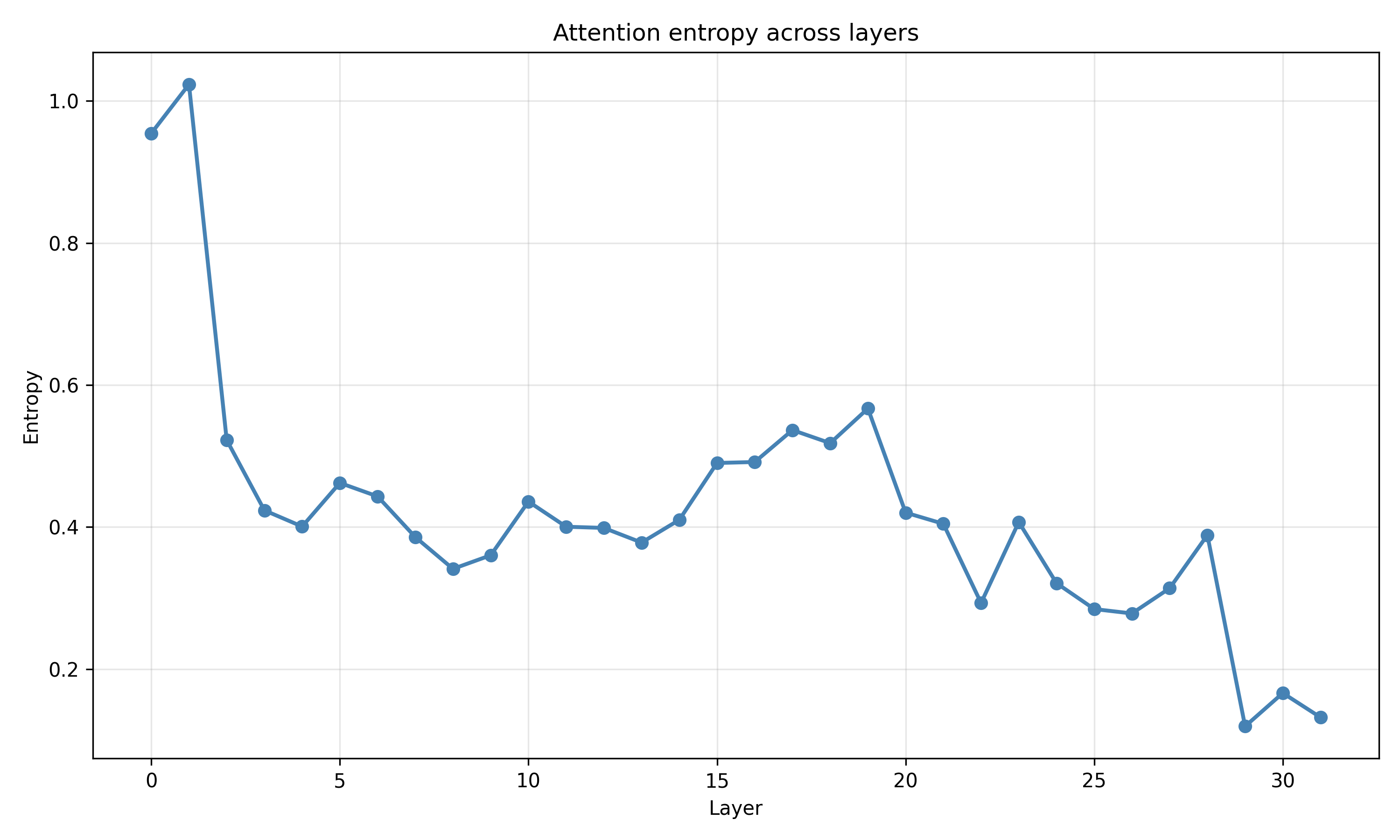}
\caption{Attention focusing}
\end{subfigure}
\caption{\textbf{Phi-2: Sharpened Bayesian geometry from curated training.}}
\label{fig:phi2}
\end{figure}
\subsection{Efficiency--Interpretability Trade-off: Llama-3.2-1B (GQA)}
\label{sec:llama-overview}
Llama-3.2-1B employs a 4:1 grouped-query attention mechanism.
\paragraph{Value manifolds.}
Mixed-domain 2D geometry (PC1=18.5\%, PC2=14.8\%); mathematics-only collapse
recovered.
\paragraph{Key orthogonality.}
Moderate: 0.15--0.18; weaker than Pythia/Phi-2 but 2$\times$ better than random.
\paragraph{Attention focusing.}
31\% entropy reduction, consistent with KV-sharing constraints.
\begin{figure}[t]
\centering
\begin{subfigure}[b]{0.32\textwidth}
\includegraphics[width=\textwidth]{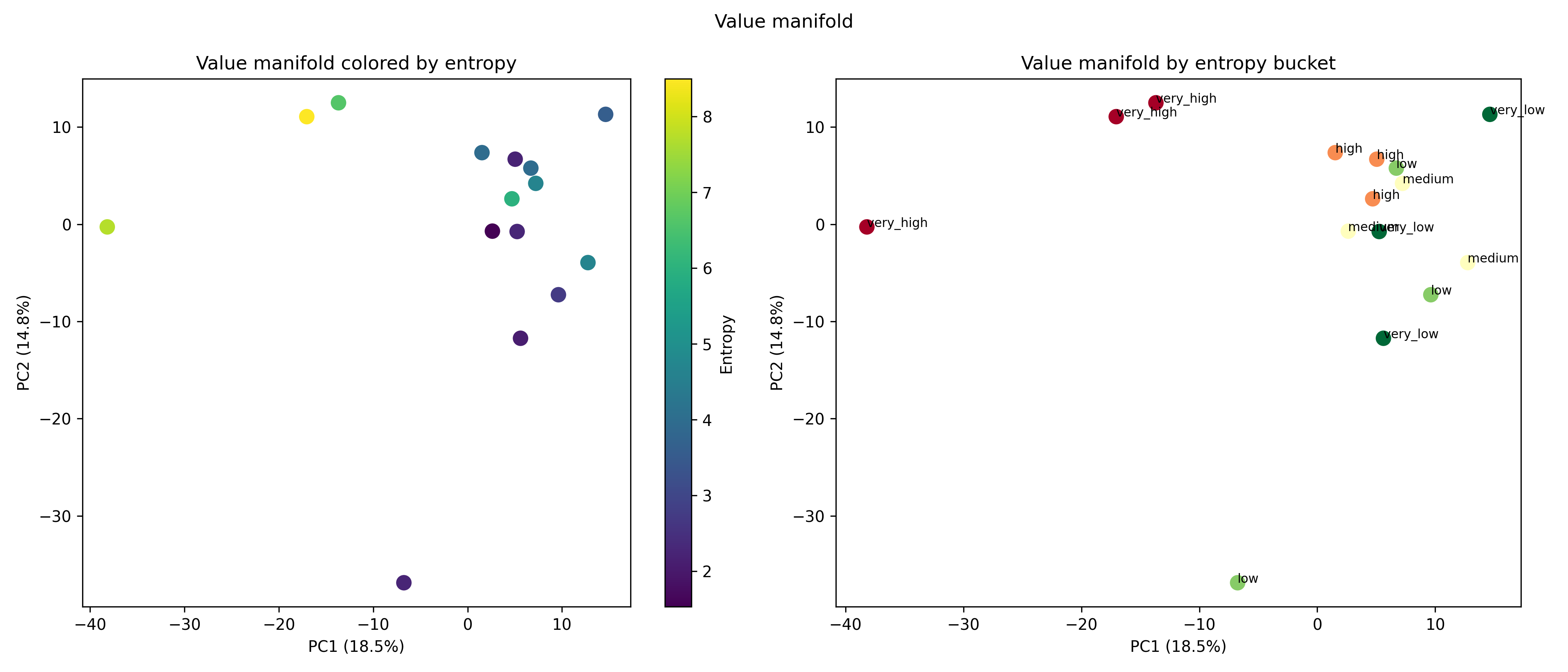}
\caption{Value manifold}
\end{subfigure}
\hfill
\begin{subfigure}[b]{0.32\textwidth}
\includegraphics[width=\textwidth]{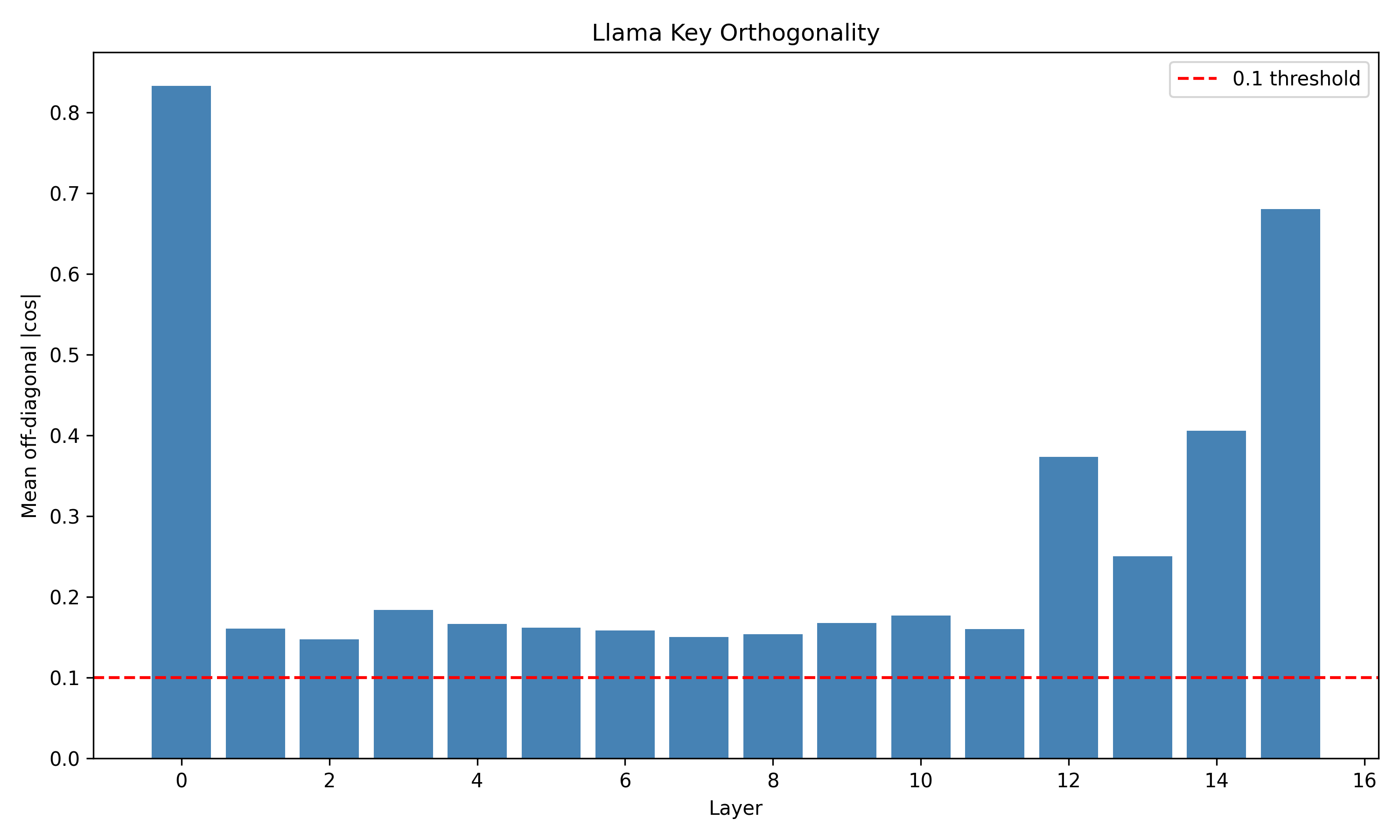}
\caption{Key orthogonality}
\end{subfigure}
\hfill
\begin{subfigure}[b]{0.32\textwidth}
\includegraphics[width=\textwidth]{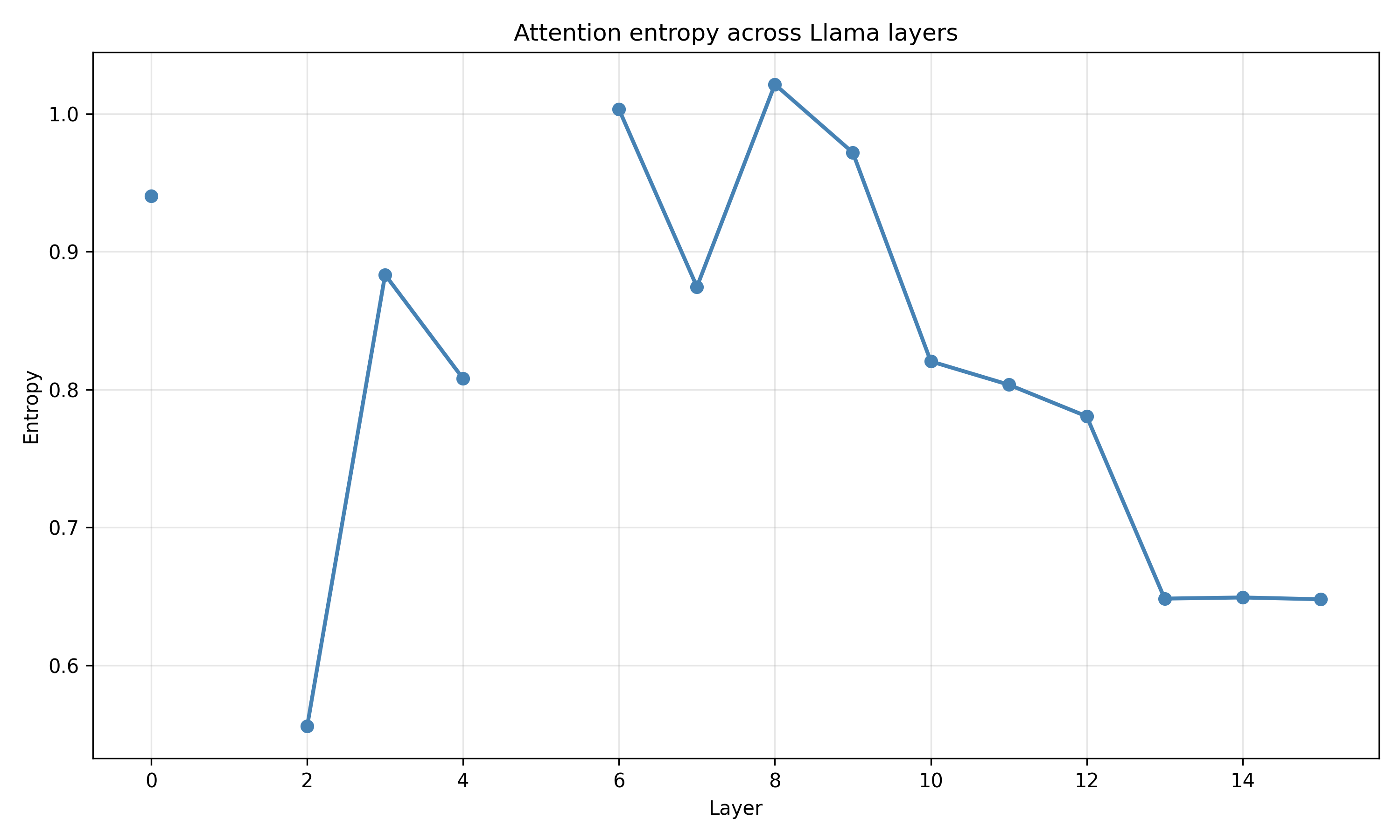}
\caption{Attention focusing}
\end{subfigure}
\caption{\textbf{Llama-3.2-1B: Bayesian structure with GQA efficiency trade-offs.}}
\label{fig:llama}
\end{figure}
\subsection{Scaling Within a Family: Pythia-12B}
\label{sec:pythia12-overview}
\paragraph{Value manifolds.}
Mixed-domain geometry becomes multi-lobed (PC1+PC2 = 19\%), but mathematics-only prompts recover a near-1D manifold (PC1+PC2 $\approx$ 0.90). Pythia-12B shows markedly different mixed-domain geometry than Pythia-410M (19\% vs 99.7\%), suggesting scale increases representational distribution. Under domain restriction both converge to similar collapse (${\sim}$90\%), indicating the geometric substrate is preserved but more distributed at scale.
\paragraph{Key orthogonality.}
Strong early (0.048--0.055), gradually decreasing with depth.
\paragraph{Attention focusing.}
Early collapse, mid-layer mixing, late refinement.
\begin{figure}[t]
\centering
\begin{subfigure}[b]{0.32\textwidth}
\includegraphics[width=\textwidth]{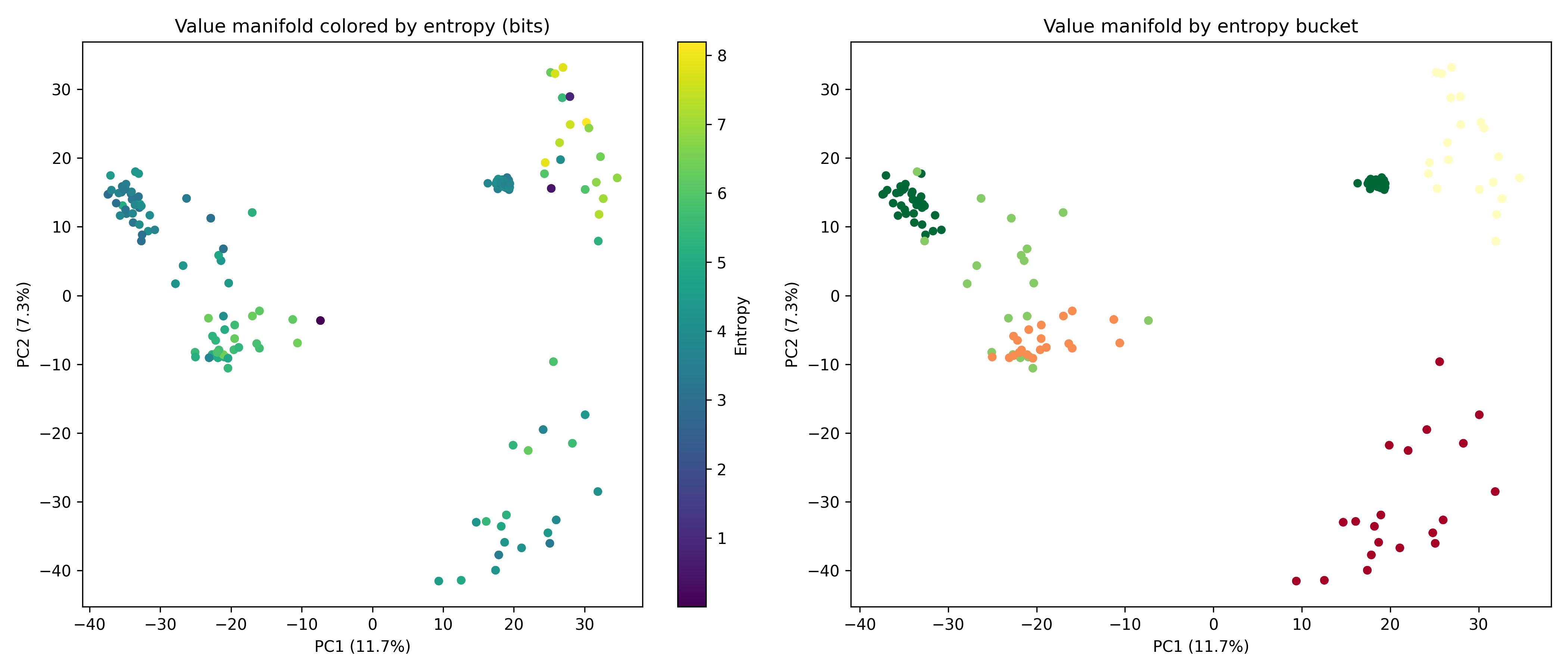}
\caption{Value manifold}
\end{subfigure}
\hfill
\begin{subfigure}[b]{0.32\textwidth}
\includegraphics[width=\textwidth]{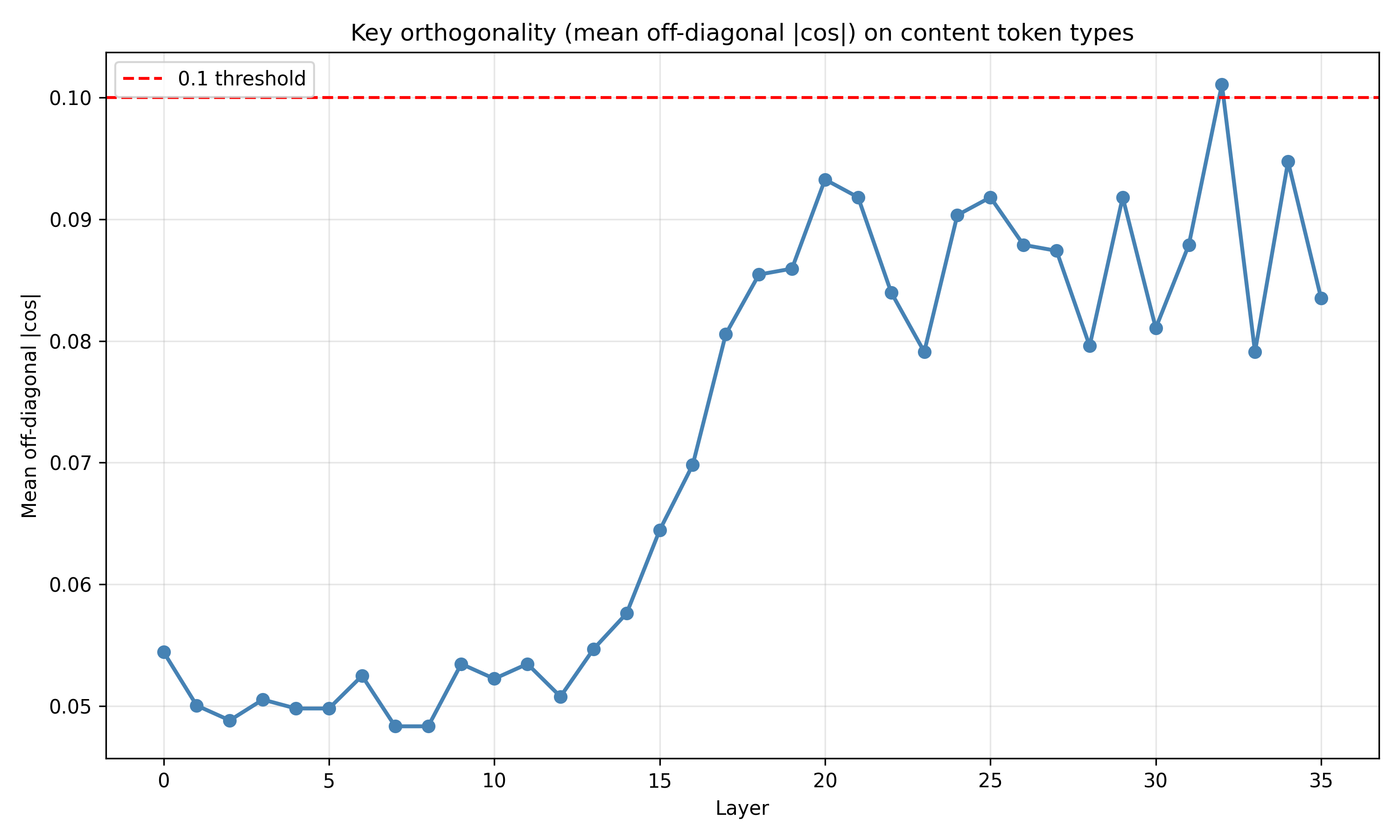}
\caption{Key orthogonality}
\end{subfigure}
\hfill
\begin{subfigure}[b]{0.32\textwidth}
\includegraphics[width=\textwidth]{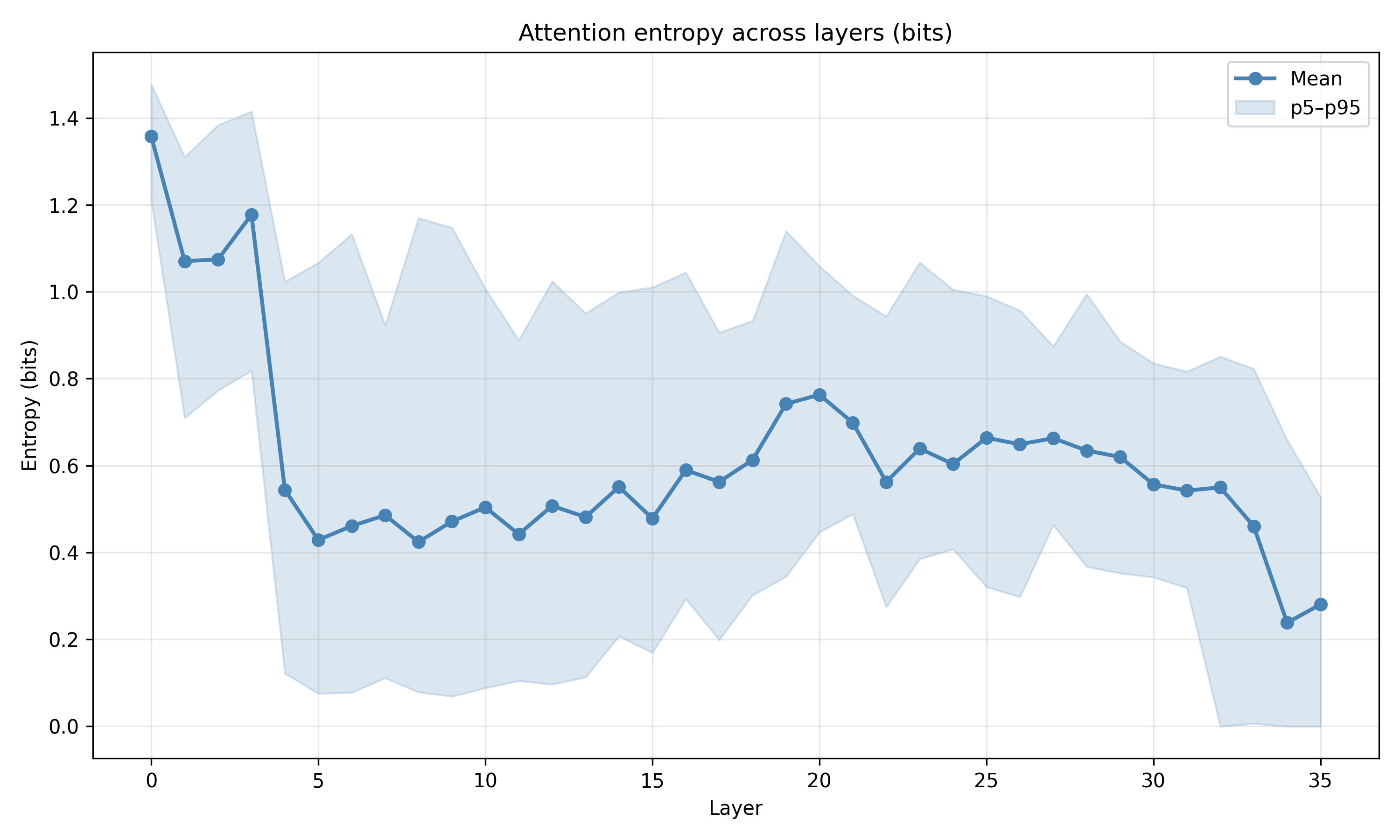}
\caption{Attention focusing}
\end{subfigure}
\caption{\textbf{Pythia-12B: Bayesian geometry at larger scale.}}
\label{fig:pythia12}
\end{figure}
\subsection{Boundary Case: The Mistral Family}
\label{sec:mistral}
\begin{figure}[t]
    \centering
    \begin{subfigure}[b]{0.32\linewidth}
        \centering
        \includegraphics[width=\linewidth]{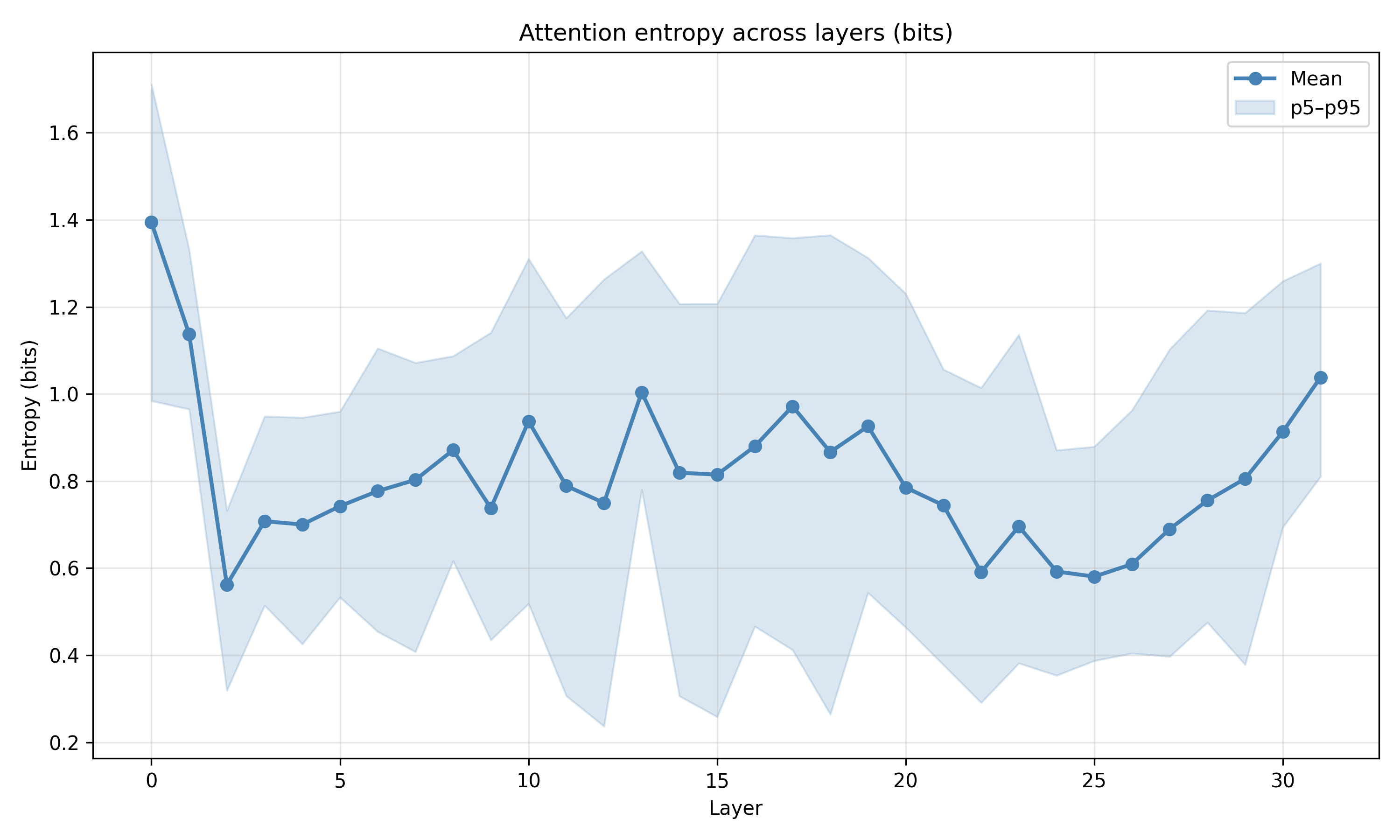}
        \caption{Mistral-7B (base)}
    \end{subfigure}
    \hfill
    \begin{subfigure}[b]{0.32\linewidth}
        \centering
        \includegraphics[width=\linewidth]{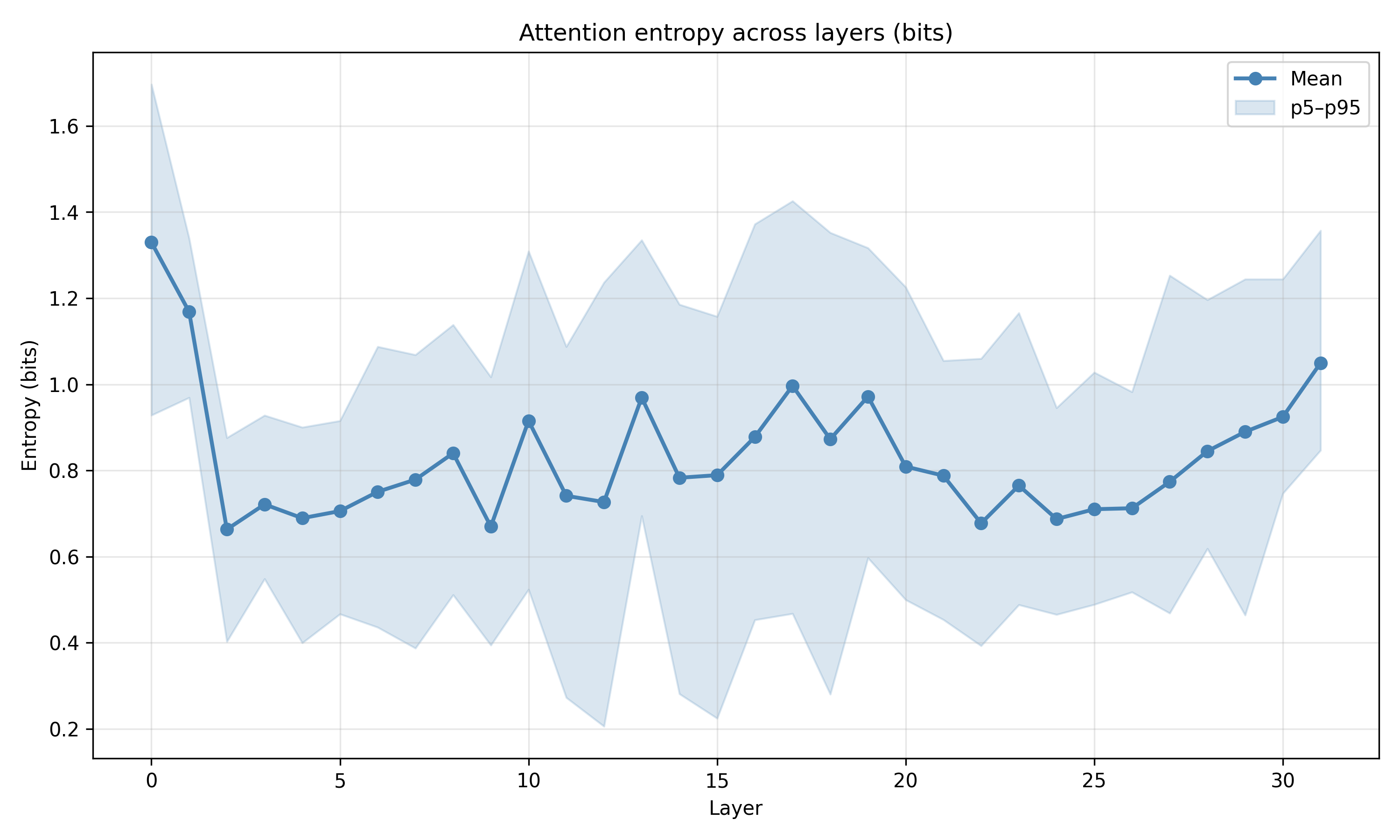}
        \caption{Mistral-7B-Instruct}
    \end{subfigure}
    \hfill
    \begin{subfigure}[b]{0.32\linewidth}
        \centering
        \includegraphics[width=\linewidth]{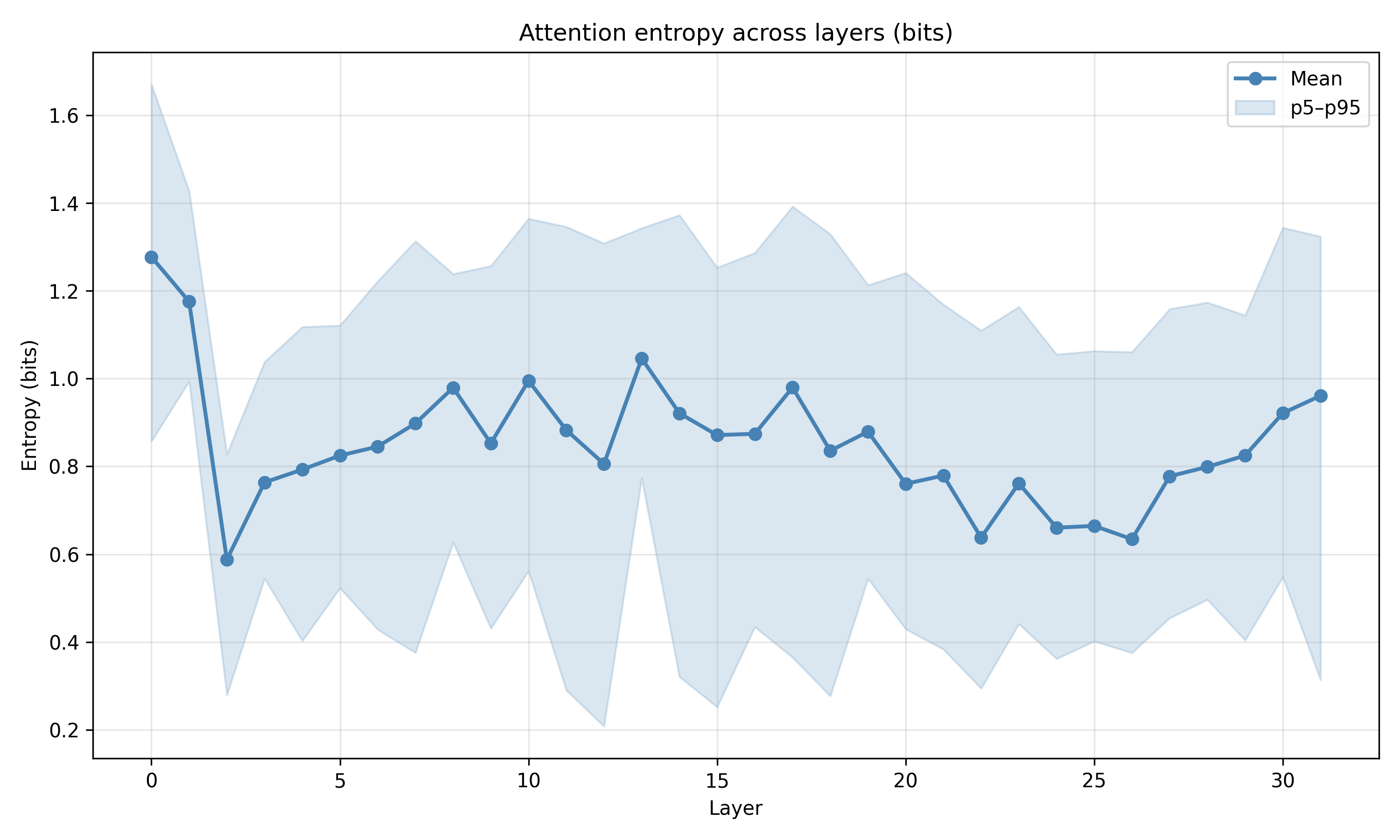}
        \caption{Mixtral-8$\times$7B}
    \end{subfigure}
    \caption{\textbf{Attenuated dynamic focusing in Mistral-style architectures.}
    Attention entropy as a function of layer depth for three variants of the
    Mistral family. Unlike models with full-sequence multi-head attention,
    entropy decreases only modestly ($20\%$--$30\%$) and often non-monotonically,
    reflecting weakened dynamic routing due to (\emph{i}) sliding-window
    attention, which prevents global evidence accumulation, and (\emph{ii})
    mixture-of-experts routing, which fragments updates across experts.
    Despite reduced focusing dynamics, the \emph{static} value-space geometry
    (\Cref{sec:domain-restriction,sec:sula}) remains intact,
    illustrating a dissociation between representational invariants and the
    mechanisms that refine them during inference.}
    \label{fig:mistral}
\end{figure}

The Mistral family provides an illuminating boundary condition for Bayesian geometry. Across all three
variants we evaluate - Mistral-7B-Base, Mistral-7B-Instruct, and the Mixtral-8$\times$7B MoE - we find
that the \emph{static} geometric signatures (value manifolds and key orthogonality) remain clean and
consistent, while the \emph{dynamic} signature (progressive attention focusing) is substantially weakened or
noisy. This dissociation reveals how architectural constraints modulate the expression of Bayesian
computations without eliminating the underlying representational substrate.
\paragraph{Static geometry persists.}
All Mistral variants exhibit low-dimensional value manifolds under mixed-domain prompts and recover
wind-tunnel-style 1D collapse under mathematics-only prompts (PC$_1{+}$PC$_2 \approx 80\%$--$90\%$).
Key orthogonality is likewise sharp: early and mid layers show mean off-diagonal cosine values near
$0.05$--$0.06$, well below both Gaussian and initialization baselines. These results indicate that the
hypothesis-frame structure and entropy-ordered manifold discovered in Papers~I--II persist in
Mistral architectures.
\paragraph{Dynamic focusing is attenuated.}
In contrast, attention entropy decreases only modestly (typically $20\%$--$30\%$) and often
non-monotonically across layers (\Cref{fig:mistral}). This stands in sharp contrast to the
binding$\rightarrow$elimination$\rightarrow$refinement trajectory observed in full-sequence MHA (Sections
5.3--5.5). The weakened focusing reflects architectural constraints:
\begin{itemize}
    \item \textbf{Sliding-window attention} restricts global routing, preventing heads from accumulating
    evidence across the entire prompt.
    \item \textbf{Mixture-of-experts (MoE) routing} fragments updates across experts, further reducing the
    coherence of evidence aggregation.
\end{itemize}
These factors disrupt the \emph{dynamic refinement} of posterior uncertainty while leaving the \emph{static}
representational geometry intact.
\paragraph{Interpretation without circularity.}
It is tempting to interpret weak focusing as a failure of Bayesian inference, but this would be circular:
strong progressive focusing is a \emph{sufficient} mechanism for Bayesian updating in full-sequence MHA, not
a \emph{necessary} one for all architectures. The Mistral models demonstrate that:
\begin{enumerate}
    \item the \emph{representational frame} for Bayesian inference (orthogonal keys + value manifold) remains
    fully intact,
    \item while the \emph{mechanism of evidence refinement} (progressive focusing) depends sensitively on global
    routing capacity.
\end{enumerate}
This pattern matches the frame--precision dissociation predicted in Paper~II: attention patterns (the frame)
stabilize early and robustly under training, whereas the precision of posterior refinement is sensitive to
architectural bandwidth and routing design.
\paragraph{Conclusion.}
The Mistral family should therefore be viewed as a boundary case that reveals how architectural constraints selectively modulate \emph{dynamic} Bayesian computation, rather than as a counterexample to Bayesian geometry. Static geometric structure---entropy-ordered manifolds and hypothesis frames---persists across
all Mistral variants, while dynamic refinement is diminished by local (sliding-window) attention. This
provides a natural explanation for the observed behavior and connects directly to the theoretical predictions
of Paper~II.
\subsection{Compressed Routing as a Boundary: The DeepSeek and Qwen Families}
\label{sec:mla-boundary}
To probe how far our geometric account extends, we evaluate two further model families: \textbf{Qwen2.5} (7B and 14B), which represents a second large GQA training regime, and \textbf{DeepSeek-V2-Lite} (16B parameters, 2.4B active), which uses \emph{Multi-head Latent Attention} (MLA)---a radically different routing mechanism in which keys and values are produced through a low-rank latent compression $c_{KV}=W_{DKV}\,x$ followed by per-head decompression $k=W_{UK}\,c_{KV}$. We additionally include \textbf{DeepSeek-LLM-7B}, a pre-MLA model from the same organization with standard MHA, as a within-developer control.
\paragraph{Methodology for MLA.}
Standard $W_K$ does not exist in MLA; we therefore extract the \emph{effective} content-key matrix $W_K^{\text{eff}}=(W_{UK}\,W_{DKV})^{\top}\in\mathbb{R}^{d_{\text{model}}\times n_h\,d_h^C}$ and apply the same orthogonality protocol as for standard MHA. Because $W_K^{\text{eff}}$ has rank at most $r=\mathrm{kv\_lora\_rank}=512$ (vs.\ $d_{\text{model}}=2048$), the appropriate random baseline is the rank-matched value $\sqrt{2/(\pi r)}\approx 0.035$ rather than the full-rank baseline $\approx 0.018$. Value-manifold and attention-entropy protocols are unchanged.
\paragraph{Static geometry persists across all four models.}
All four models exhibit the canonical static signatures (\Cref{tab:cross-architecture}). Domain restriction collapses value manifolds toward one dimension in every case: PC$_1{+}$PC$_2$ rises from 21.1\%$\to$48.1\% (Qwen2.5-7B), 26.8\%$\to$53.3\% (Qwen2.5-14B), 20.4\%$\to$50.2\% (DeepSeek-LLM-7B), and 14.1\%$\to$39.1\% (DeepSeek-V2-Lite). Key orthogonality is uniformly sharp in the standard-attention models (0.04--0.09 in early layers), with one notable exception described below.
\paragraph{MLA produces a layer-invariant orthogonality plateau.}
DeepSeek-V2-Lite shows mean off-diagonal cosine of $0.0429$--$0.0434$ \emph{at every one of its 27 layers} (computed in fp32 to rule out quantization). The 0.005-point spread is far smaller than within-layer head-to-head variation in standard MHA. Two factors explain this: (i)~the rank-512 latent bottleneck constrains all keys to a fixed-dimension subspace regardless of layer, and (ii)~within that subspace the measured value lies only $\sim 22\%$ above the rank-matched random baseline ($\sqrt{2/(\pi r)}\approx 0.035$). Standard MHA models, by contrast, achieve roughly $2$--$10\times$ improvement over their full-rank baselines.
\paragraph{The decoupled RoPE channel is also near-random.}
A natural hypothesis is that hypothesis discrimination migrates to the decoupled RoPE channel $k_R=W_{KR}\,x$, since RoPE keys are applied per token and avoid the latent bottleneck. We test this by extracting $W_{KR}\in\mathbb{R}^{q_R\times d_{\text{model}}}$ (rows $r{:}r{+}q_R$ of \texttt{kv\_a\_proj\_with\_mqa}) and applying the same off-diagonal-cosine protocol. Across all 27 layers, mean RoPE-channel cosine is $0.1010$--$0.1021$---essentially indistinguishable from the random Gaussian baseline of $\sqrt{2/(\pi\cdot 64)}\approx 0.0997$ for the 64-dimensional RoPE space. Standard deviation across layers is $0.0003$. Both channels of MLA are therefore near-isotropic in static directional terms; whatever per-prompt discrimination the model performs is achieved through learned \emph{coefficients} on these fixed bases (i.e., via the multiplicative interaction with $c_{KV}$ during decompression), not through directional sharpening of the projection matrices themselves. The static-orthogonality signature, central to standard-MHA Bayesian geometry, simply does not transfer to MLA.
\paragraph{Dynamic focusing is absent in MLA.}
Layerwise attention entropy reduction in DeepSeek-V2-Lite is just \emph{4.6\%} mixed and 4.5\% math, an order of magnitude weaker than any other architecture in our study. Even Mistral's sliding-window and MoE variants reach 20--30\%. The same-organization control DeepSeek-LLM-7B (standard MHA) exhibits 51\% reduction, ruling out training pipeline or corpus effects: the gap is attributable to MLA itself. We interpret this as the strongest instance of the frame--precision dissociation predicted in Paper~II: MLA's compression preserves the representational frame (value manifold, low-rank hypothesis frame) but eliminates the per-layer routing capacity that drives progressive posterior refinement in standard MHA.
\paragraph{Attention focusing decreases with depth in the Qwen family.}
Qwen2.5-7B (28 layers) achieves $\downarrow$65\% entropy reduction, while Qwen2.5-14B (48 layers) achieves only $\downarrow$24\%. Both models share GQA and RoPE; the difference is depth and KV-sharing ratio (7:1 vs.\ 5:1). This contrasts with the Pythia family, where focusing \emph{increases} from 410M to 12B---suggesting that scale interacts with architecture to determine whether refinement concentrates in a few layers or distributes across many. The static signatures (value manifold collapse, key orthogonality) are stable in both Qwen sizes, consistent with the broader pattern that representation is universal but mechanism is architecture-dependent.
\begin{figure}[t]
    \centering
    \begin{subfigure}[b]{0.48\linewidth}
        \centering
        \includegraphics[width=\linewidth]{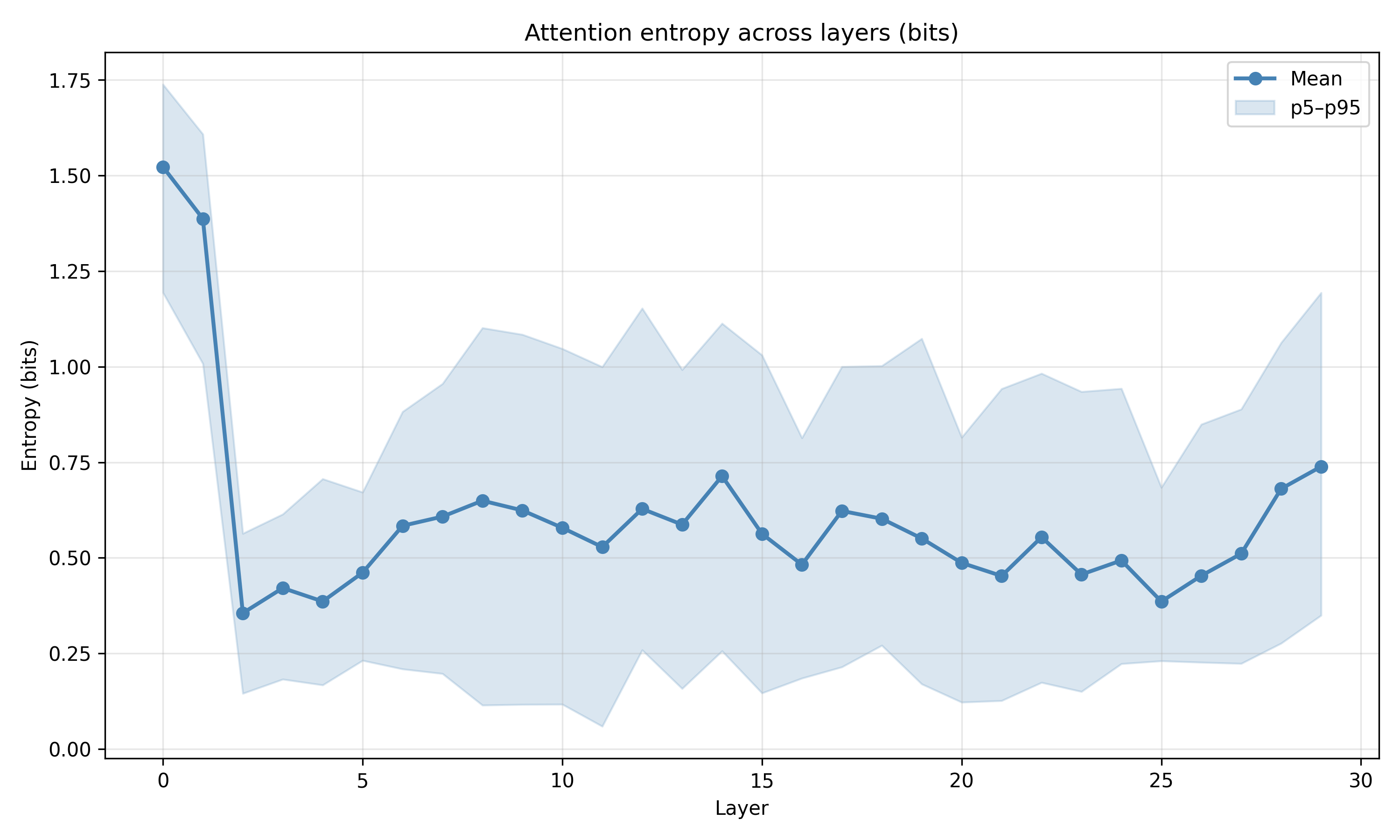}
        \caption{DeepSeek-LLM-7B (standard MHA): $\downarrow$51\% entropy reduction.}
    \end{subfigure}
    \hfill
    \begin{subfigure}[b]{0.48\linewidth}
        \centering
        \includegraphics[width=\linewidth]{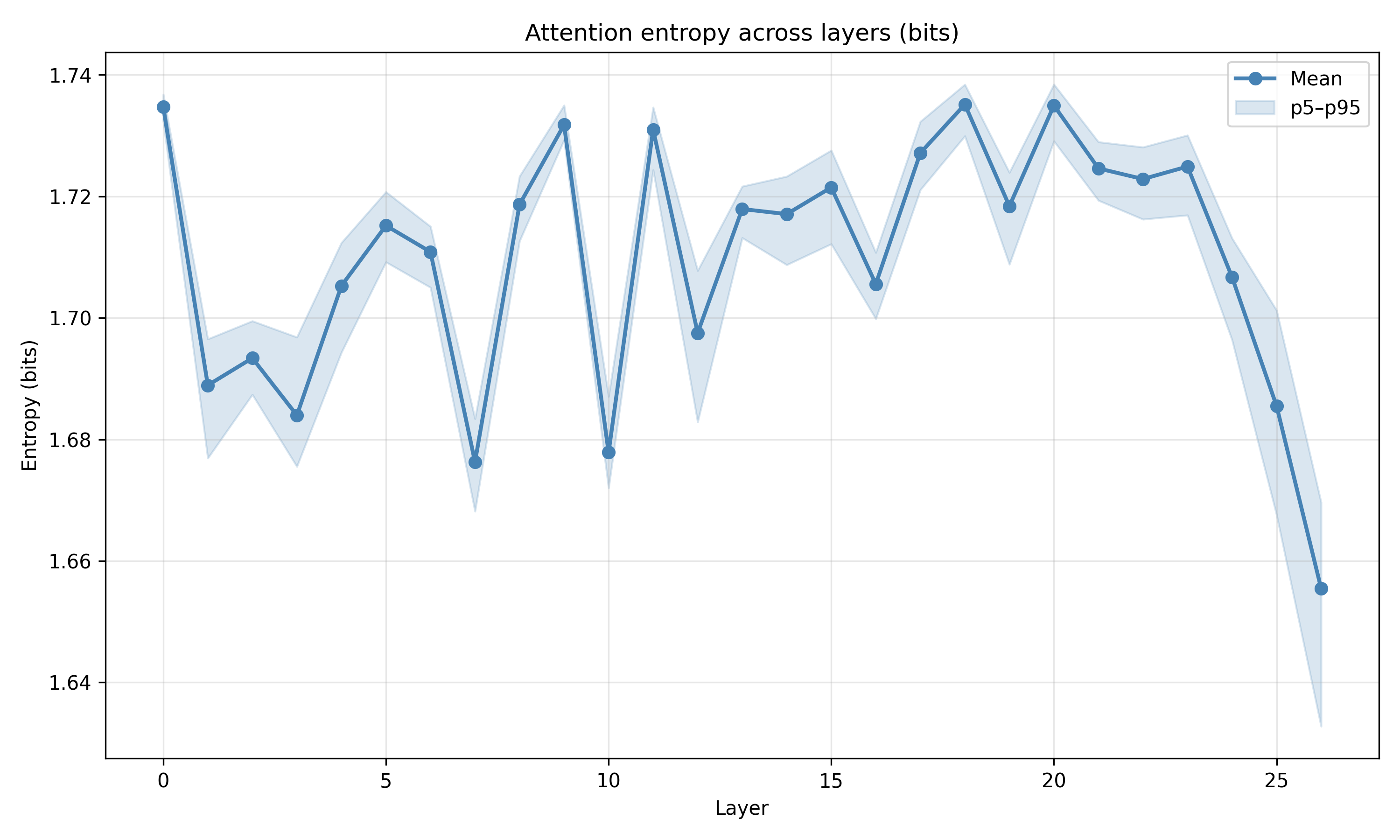}
        \caption{DeepSeek-V2-Lite (MLA+MoE): $\downarrow$5\% (essentially flat).}
    \end{subfigure}
    \\[6pt]
    \begin{subfigure}[b]{0.48\linewidth}
        \centering
        \includegraphics[width=\linewidth]{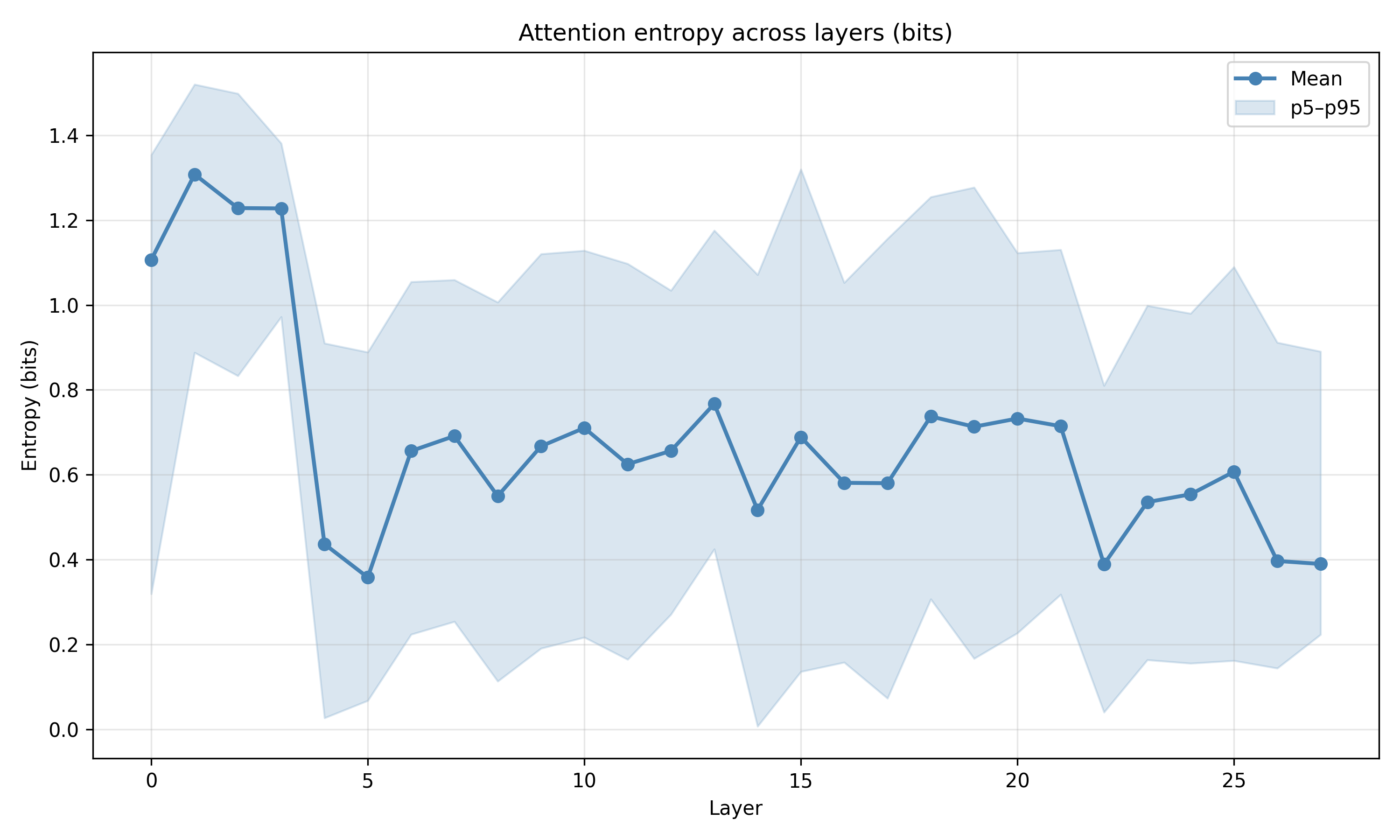}
        \caption{Qwen2.5-7B (GQA, 28 layers): $\downarrow$65\%.}
    \end{subfigure}
    \hfill
    \begin{subfigure}[b]{0.48\linewidth}
        \centering
        \includegraphics[width=\linewidth]{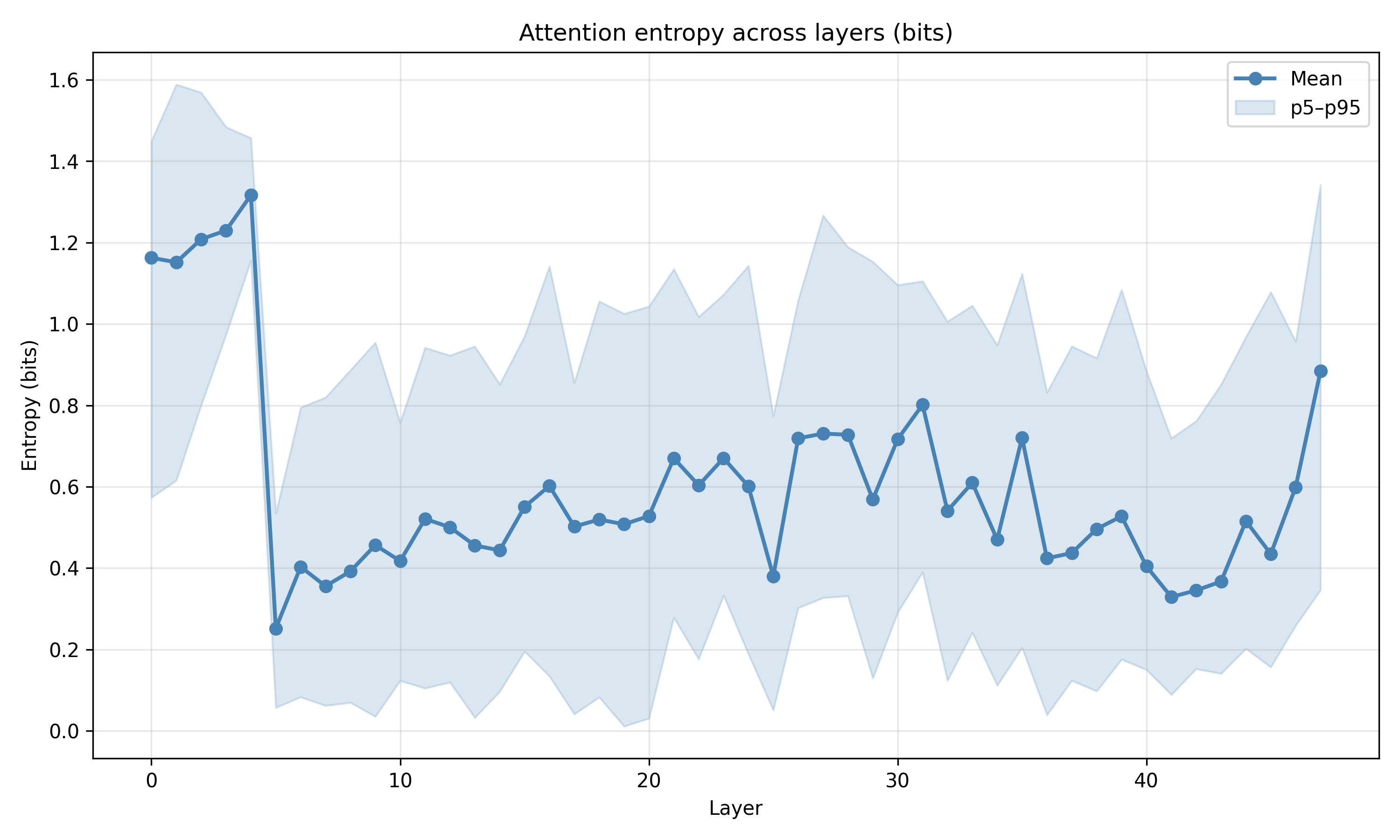}
        \caption{Qwen2.5-14B (GQA, 48 layers): $\downarrow$24\%.}
    \end{subfigure}
    \caption{\textbf{Attention focusing across compressed-routing and GQA architectures.}
    Top: a same-organization comparison isolates the MLA effect---DeepSeek-LLM-7B (standard MHA) shows the canonical layer-wise sharpening; DeepSeek-V2-Lite (MLA+MoE) is essentially flat across all 27 layers despite identical training scale and corpus.
    Bottom: within the Qwen2.5 family, attention focusing weakens substantially when moving from 28 to 48 layers, suggesting that depth and KV-sharing ratio interact to redistribute the dynamic refinement across more layers, lowering per-layer reduction.
    Static signatures (value manifolds, key orthogonality) remain intact in all four cases.}
    \label{fig:mla-attn}
\end{figure}
\paragraph{Implication for the trilogy's thesis.}
Paper~III's discussion (\Cref{sec:discussion}) frames \emph{content-based value routing} as the essential ingredient for Bayesian geometry. The DeepSeek-V2-Lite results sharpen this claim along two dimensions. \emph{Architecture-time:} the static value manifold and its domain-restriction collapse survive MLA's rank-512 compression---the representational scaffolding is preserved. \emph{Inference-time:} the SULA experiment (\Cref{sec:sula}) shows that this scaffolding is not actually \emph{used} for Bayesian updating---PC$_1$ neither aligns with the analytic posterior ($\rho\approx -0.03$) nor moves with $k$ ($\rho\approx -0.01$), and predictive-entropy MAE is the worst we observe (0.59 bits, vs.\ 0.23 for Qwen2.5-7B). Progressive attention focusing, similarly, is absent ($\downarrow$5\%). Compressed-routing MLA is therefore a clean dissociation: the trilogy's static signatures are present, the dynamic mechanism that engages them at inference time is not. We read this as evidence that ``content-based value routing'' is necessary but not sufficient---when the routing channel is forced through a low-rank bottleneck, the static substrate forms during pretraining but cannot be recruited for in-context Bayesian updating.
\subsection{Cross-Architecture Synthesis}
\label{sec:cross-architecture}
\begin{table}[t]
\centering
\caption{Bayesian geometric signatures across architectures. Value-manifold dimensionality reports PC1+PC2 under mixed-domain and domain-restricted (mathematics) prompts. Key orthogonality shows mean off-diagonal cosine in early layers. Attention focusing reports entropy reduction from first to final layer.}
\label{tab:cross-architecture}
\small
\begin{tabular}{llccccc}
\toprule
\textbf{Model} & \textbf{Arch} & \textbf{Training} & \multicolumn{2}{c}{\textbf{Value Manifold}} & \textbf{Key Orthog} & \textbf{Attn Focus} \\
 & & & Mixed & Math & (early) & \\
\midrule
Pythia-410M & MHA & Pile & 99.7\% & 99.9\% & 0.11--0.13 & $\downarrow$82\% \\
Phi-2 & MHA & Curated & 60.6\% & 63.5\% & 0.034--0.051 & $\downarrow$86\% \\
Pythia-12B & MHA & Pile & $\sim$19\% & $\sim$90\% & 0.05--0.10 & non-monotone \\
DeepSeek-LLM-7B & MHA & Web & 20.4\% & 50.2\% & 0.053--0.078 & $\downarrow$51\% \\
Llama-3.2-1B & GQA(4:1) & Web & 51.4\% & 73.6\% & 0.15--0.18 & $\downarrow$31\% \\
Qwen2.5-7B & GQA(7:1) & Web & 21.1\% & 48.1\% & 0.063--0.093 & $\downarrow$65\% \\
Qwen2.5-14B & GQA(5:1) & Web & 26.8\% & 53.3\% & 0.045--0.061 & $\downarrow$24\% \\
Mistral-7B & GQA+SW & Web & 15--20\% & $\sim$80\% & 0.05--0.06 & 20--30\% \\
DeepSeek-V2-Lite & \textbf{MLA}+MoE & Web & 14.1\% & 39.1\% & 0.043 (flat) & $\downarrow$5\% \\
\midrule
Wind-Tunnel & MHA & Synthetic & --- & 84--90\% & 0.09--0.12 & $\downarrow$85--90\% \\
\bottomrule
\end{tabular}
\vspace{0.5em}
\begin{flushleft}
\footnotesize
\textbf{Notes:} Value manifold dimensionality varies substantially across architectures under mixed-domain prompts. Pythia-410M shows near-complete collapse regardless of domain; Llama-3.2-1B, Qwen2.5-7B, and DeepSeek-V2-Lite all show clear domain-restriction effects (PC$_1{+}$PC$_2$ approximately doubling under math-only prompts). All MHA and GQA models achieve key orthogonality 2--10$\times$ better than random Gaussian baselines ($\approx$0.11 for $d_k=64$). DeepSeek-V2-Lite's MLA architecture compresses keys through a rank-512 bottleneck, yielding nearly layer-invariant orthogonality at $\approx 0.043$, only mildly sharper than the rank-matched random baseline ($\approx 0.035$). Attention focusing depends strongly on architecture: full-sequence MHA achieves strong progressive reduction (51--86\%); GQA shows mixed results (24--65\%, decreasing with model depth in the Qwen family); sliding-window variants show weaker or non-monotone patterns (20--30\%); and MLA exhibits essentially no progressive focusing ($<$5\%), the strongest static/dynamic dissociation observed.
\end{flushleft}
\end{table}
\Cref{tab:cross-architecture} summarizes the unified picture across nine production models spanning MHA, GQA, sliding-window, MoE, and MLA architectures:
\begin{itemize}
\item \textbf{Value manifolds:} all models exhibit low-dimensional value
geometry. Under mixed-domain prompts, $\mathrm{PC}_1{+}\mathrm{PC}_2$
varies substantially across architectures---from $\sim$14\% in DeepSeek-V2-Lite and $\sim$15--20\% in Mistral to $\sim$51\% in Llama and $\sim$61\% in Phi-2, with Pythia-410M as a notable
outlier whose manifold is nearly collapsed even in the mixed setting
($\sim$100\%). Under mathematics-only prompts, all models exhibit substantial collapse, with most moving into the
40--95\% range, approaching the one-dimensional structure of the
wind-tunnel tasks. The domain-restriction effect (math vs.\ mixed PC$_1{+}$PC$_2$ ratio) is most pronounced in Llama, Qwen, and DeepSeek variants and minimal in models with intrinsically collapsed manifolds (Pythia-410M, Phi-2).
\item \textbf{Key orthogonality:} mean off-diagonal cosine is consistently
2--10$\times$ lower (better) than random baselines or initialization in all standard-attention models,
indicating robust hypothesis-frame structure. MLA is a special case (\Cref{sec:mla-boundary}): the rank-512 latent compression flattens the orthogonality signal across layers and reduces the gap to the rank-matched random baseline.
\item \textbf{Attention focusing:} layerwise entropy reduction varies
systematically by architecture--strong in full-sequence MHA (51--86\%), moderate to weak in
GQA (24--65\%, with depth-dependent attenuation in the Qwen family), and modest and often non-monotone in sliding-window and MoE
variants (20--30\%). MLA exhibits essentially zero progressive focusing ($<$5\%), the strongest dissociation between intact static geometry and absent dynamic refinement we observe.
\end{itemize}
As summarized in \Cref{tab:cross-architecture}, all three geometric signatures show
quantitative continuity from synthetic wind-tunnel tasks to production models.
\paragraph{Static vs dynamic geometry.}
The stable invariants across every architecture - including Mistral - are:
(1) entropy-ordered value manifolds, and (2) orthogonal key frames.  
Dynamic focusing is architecture-dependent, requiring global routing capacity.
This separation is the central representational--computational split predicted in
Paper~II.
\begin{figure}[t]
    \centering
    \includegraphics[width=0.92\linewidth]{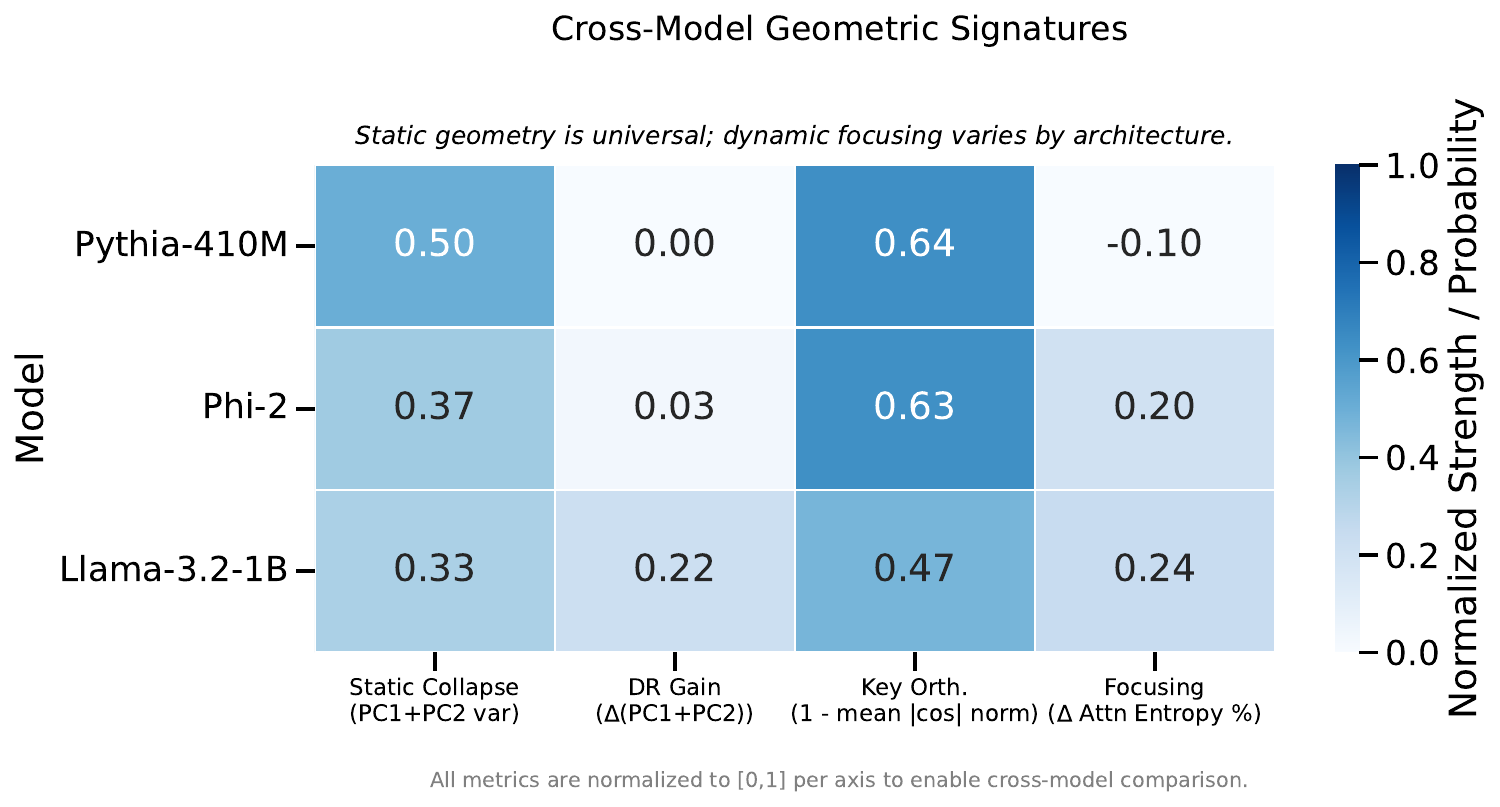}
    \caption{\textbf{Cross-model geometric signatures.}
    Normalized comparison of four geometric metrics across three model families. 
    \emph{Left:} Static signatures---value manifold collapse (PC$_1{+}$PC$_2$) and
    domain-restriction gain---are consistently present, indicating that all models 
    internalize a low-dimensional representation of hypothesis space.
    \emph{Right:} Dynamic signatures---key orthogonality and attention focusing---vary
    substantially by architecture: Llama-3.2-1B exhibits strong runtime refinement,
    whereas Pythia-410M relies on pre-orthogonalized hypothesis frames and shows 
    weak focusing. This dissociation suggests that representation is universal, 
    but the mechanism that refines it during inference is architectural.}
    \label{fig:cross_model_summary}
\end{figure}
\section{Analysis and Key Findings}
\label{sec:findings}
Our cross-architecture validation reveals systematic patterns in how Bayesian geometric structures scale from wind tunnels to production models.
\subsection{Universal core mechanisms}
\label{sec:universal}
Across the standard MHA and GQA models (Pythia, Phi-2, and Llama), we observe the full set of Bayesian
geometric signatures predicted by the wind tunnel experiments: value manifolds, key orthogonality, and
layerwise attention focusing. In the Mistral family (\Cref{sec:mistral}), the static signatures (value
manifolds and key orthogonality) persist, while the dynamic signature (monotone attention focusing) is
weakened or absent due to architectural and training-objective differences. This suggests a natural split
between \emph{universal static structure} and \emph{architecture-dependent dynamics}.
\textbf{Value manifolds (static).} All models, including the Mistral variants, show low-dimensional value
structure, with PC1+PC2 between roughly 16\% and 84\%, far above random baselines. Entropy ordering
persists in every model evaluated: prompts with higher next-token entropy occupy systematically different
regions of the manifold than low-entropy prompts.
\textbf{Key orthogonality (static).} All models learn hypothesis-frame structure, with mean off-diagonal
cosine between 0.034 and 0.18, consistently 2 - 10 times better than random initialization (0.40 - 0.45).
Mistral models show the same early-layer emergence and late-layer collapse of orthogonality as standard MHA.
\textbf{Attention focusing (dynamic).} Standard MHA and GQA models exhibit a clear layerwise entropy
decrease (31--86\%), matching the binding $\rightarrow$ elimination $\rightarrow$ refinement pattern seen in wind
tunnels. In the Mistral models, this dynamic signature is weak or absent (\Cref{sec:mistral}), consistent
with architectural constraints (sliding-window attention limiting global context access) and the effects of downstream
alignment. The 50+ percentage point gap between MHA models (82--86\%) and GQA/sliding-window models (20--31\%) indicates that architectural constraints can substantially limit dynamic Bayesian refinement even when static geometry is preserved.
The ICL experiment connects the controlled wind-tunnel setting to real-time
inference in production models. In the wind tunnels, Bayesian structure
emerges from the training objective alone: the model learns to encode
posterior uncertainty along a one-dimensional manifold. The ICL setting
demonstrates that the same mechanism is active \emph{during inference}: when
the model is given explicit symbolic evidence inside the prompt, its value
representations move along the same manifold direction that encodes posterior
entropy, and the degree of movement is proportional to the amount of
evidence. In this way, the ICL results show that transformers not only embody
geometry consistent with Bayesian inference as a representational primitive, but also execute updates
correlated with Bayesian posteriors when the prompt supplies usable likelihood information.
\textbf{Conclusion.} The static components of the Bayesian geometry (value manifolds and hypothesis frames)
are universal across architectures, while strong progressive attention focusing depends on global routing
capacity and training regime.
\subsection{Training Data Quality Enhances Clarity}
Geometric clarity correlates strongly with training data quality:
\begin{itemize}
\item \textbf{Phi-2} (curated textbooks/code): Orthogonality 0.034 - 0.051, focusing 86\%
\item \textbf{Pythia} (Pile corpus, diverse): Orthogonality 0.11 - 0.13, focusing 82\%
\item \textbf{Llama} (web-scale): Orthogonality 0.15 - 0.18, focusing 31\%
\end{itemize}
High-quality, consistent training examples enable gradient dynamics to sculpt sharper hypothesis frames (better orthogonality) and stronger attention pathways (better focusing).
\textbf{Implication for training}: Early training on curated data may establish cleaner geometric scaffolding that persists through subsequent web-scale training. Curriculum learning - progressing from structured to diverse data - could enhance both interpretability and reliability.
\paragraph{Causal resolution of the Bayesian manifold.}
Our final experiment simultaneously ablates the entropy-aligned value direction
across all identified ``Bayesian layers'' (8, 12, 16, 20, 23) in Pythia-410M.
If this axis encoded a causal bottleneck for Bayesian inference, removing it at
every layer should degrade and ultimately collapse SULA behavior. Instead, we
observe the opposite dissociation: (i) the multi-layer ablation destroys the
entropy geometry (axis--entropy correlation drops from $0.27$ to $0.07$), yet
(ii) SULA calibration remains intact, with \textsc{MAE} and correlation to the
Bayesian posterior changing by less than $1\%$, and (iii) a matched random-axis
ablation \emph{improves} calibration metrics despite preserving the original
geometry. These results rule out both single-axis and multi-axis bottleneck
hypotheses and show that the entropy manifold is a \emph{representational
trace} of inference rather than the substrate that performs it.
\subsection{Architectural Trade-offs: Efficiency vs Interpretability}
Grouped-query attention demonstrates clear efficiency-interpretability trade-offs:
\begin{itemize}
\item \textbf{Efficiency gains}: 4$\times$--- reduction in KV cache size, 4$\times$--- faster inference
\item \textbf{Clarity costs}: 50\% weaker orthogonality, 62\% weaker focusing (31\% vs 82\%)
\item \textbf{Functional preservation}: Qualitative Bayesian structures persist
\end{itemize}
The 4:1 query-to-KV ratio forces keys and values to serve multiple query groups simultaneously, preventing the sharp specialization seen in standard MHA. However, the model compensates by distributing information across dimensions (2D manifolds, distributed attention patterns) while maintaining overall Bayesian computational structure.
The Mistral family provides a complementary case: sliding-window
attention weakens dynamic focusing even when static
geometry (orthogonality, value manifolds) remains intact.
\textbf{Implication for deployment}: GQA is suitable for production deployment where efficiency matters, but researchers studying mechanisms or building interpretability tools should prefer standard MHA for clearer geometric signatures.
\subsection{Depth Drives Richer Representations}
Value manifold dimensionality correlates with depth:
\begin{itemize}
\item \textbf{Pythia} (24 layers): 1D manifolds (PC1=84\%, PC2=4.5\%)
\item \textbf{Phi-2} (32 layers): 2D manifolds (PC1=21\%, PC2=13\%)
\item \textbf{Llama} (16 layers, but GQA): 2D manifolds (PC1=18.5\%, PC2=14.8\%)
\end{itemize}
Deeper architectures develop richer uncertainty representations requiring additional dimensions. The key point: entropy parameterization persists---the additional dimension is not noise but structured geometry enabling finer posterior modeling.
Depth appears to enable transformers to represent multimodal or high-dimensional uncertainty distributions while shallower models compress to 1D entropy coordinates.
\subsection{Layer-wise Functional Specialization}
Consistent patterns emerge across models:
\textbf{Layer 0}: Setup phase - high key similarity, moderate attention entropy. Likely initializes representations (position embeddings, token embeddings) before geometric structure emerges.
\textbf{Layers 1 - N-4}: Core computation - strong orthogonality, progressive attention focusing. These layers perform Bayesian hypothesis discrimination and evidence accumulation.
\textbf{Final 3 - 4 layers}: Collapse phase - orthogonality weakens, attention sharpens dramatically. Hypothesis frames collapse as model commits to output distribution.
The Mistral models follow the same structural pattern for value manifolds
and key orthogonality, but do not show the expected sharpening of
attention in the elimination and refinement stages. This is consistent
with constraints on global routing (sliding window) and fragmented
updates (MoE).
This functional stratification mirrors the three-stage inference process: binding context $\rightarrow$ eliminating hypotheses $\rightarrow$ refining output.
\paragraph{Bayes calibration in natural language vs.\ wind--tunnel tasks.}
Entropy calibration errors in the ICL setting (0.31--0.44 bits across models)
are considerably larger than those observed in the controlled wind--tunnel
experiments (typically below 0.05 bits). This gap is expected: natural--language
prompts introduce substantial semantic ambiguity, and production models are not
trained on the synthetic labeling distribution used in our ICL task. The key
result is therefore not the absolute calibration error but the \emph{systematic
correspondence} between value--manifold coordinates, model predictive entropy,
and the analytical Bayesian posterior as evidence accumulates. This indicates
that the same geometric substrate identified in wind--tunnel training is actively
used by production models during inference.
\subsection{Robustness and Limitations}
\label{sec:limitations}
\paragraph{What transfers robustly.}
Several geometric signatures persist across all dense, GQA, and sliding-window/MoE architectures
evaluated. First, \emph{value manifolds} are consistently low-dimensional: PC$_1{+}$PC$_2$ remains far above
random (ranging from ~15
below heuristic thresholds. Second, \emph{key orthogonality} shows the characteristic pattern predicted by
wind-tunnel analyses: sharp early-layer frames followed by gradual late-layer collapse. Third,
\emph{layerwise functional structure} - setup layers, a broad computation band, and final collapse
layers - appears in all models at the level of static geometry.
\paragraph{What varies systematically.}
Quantitative clarity depends on architecture, training data, and depth. GQA reduces the sharpness of
orthogonality and focusing; web-scale training reduces geometric contrast relative to curated data; and
deeper models develop richer or multi-lobed manifolds under mixed prompting. Dynamic attention
focusing depends most sensitively on architecture: strong in full-sequence MHA, moderate in GQA, and
weak or noisy in sliding-window and MoE variants.
\paragraph{What domain restriction isolates.}
Domain restriction functions as a natural intervention: it reduces the multiplicity of task-specific inference
modes activated by mixed prompts. When prompts come from a single coherent domain (e.g.,
mathematics), the model operates in a more homogeneous inference regime, revealing the same collapsed
1D manifold observed in wind-tunnel settings. Note that domain restriction \emph{does not} prove that
the model performs ``true-- Bayesian inference on natural language; rather, it shows that a \emph{Bayesian
geometric coordinate system} is embedded in its representation space and can be isolated when the prompt
distribution reduces task heterogeneity.
Pythia-410M departs from this pattern, exhibiting an intrinsically
low-dimensional value space (PC$_1{+}$PC$_2 \approx 99.7\%$) under
both mixed and mathematics-only prompts (\Cref{tab:domain-pca}),
suggesting that its final-layer values are effectively collapsed
regardless of domain. We attribute this to the Pile's relatively homogeneous token distribution compared to more diverse web-scale corpora; the model appears to operate in a near-single-mode regime.
\paragraph{Causal limitations.}
Our findings are correlational. Static and dynamic geometric signatures co-occur with Bayesian-like
behavior, but we do not intervene directly on the geometry to test necessity. Establishing causal roles
would require controlled manipulations - for example:
\begin{itemize}
    \item degrading or sharpening key orthogonality and measuring effects on calibration,
    \item perturbing value vectors along or orthogonal to the manifold axes,
    \item ablating heads responsible for attention refinement.
\end{itemize}
Developing such interventions without collapsing the model's function entirely remains technically
challenging. The present results show that geometry aligns with Bayesian computations, but they do not
establish that geometry is \emph{required} for these computations.
\paragraph{Open representational questions.}
The emergence of 2D or multi-lobed manifolds in deeper or larger models is not yet theoretically
understood. These structures may encode multimodal uncertainty, semantic clustering, training-set
heterogeneity, or task-mixture effects. Likewise, the interaction between positional embeddings, local
attention kernels, and geometric formation remains an open problem, especially in sliding-window or
hybrid transformer--SSM architectures.
\paragraph{Scale and architecture coverage.}
Our largest dense model is 12B parameters, and our largest MoE model is Mixtral-8$\times$7B. Although
consistent patterns appear from 410M to 12B, evaluating frontier-scale checkpoints (70B--400B) is
necessary to determine whether new geometric phenomena arise or whether multi-lobed structure
remains the dominant pattern under mixed-domain prompts.
Overall, while the geometric signatures are consistent, a complete causal and mechanistic account
of their formation - and of their architectural modulation - remains an important direction for future work.
\section{Discussion}
\label{sec:discussion}
This paper extends the geometric account of Bayesian inference developed in
Papers~I--II to production-scale language models. Three results stand out:
domain restriction collapses value manifolds to the one-dimensional geometry
characteristic of exact Bayesian wind-tunnel tasks; transformers navigate this
geometry during inference in the SULA experiment; and static geometric
signatures - entropy-ordered value manifolds and orthogonal key frames - appear
across all architectures evaluated, including GQA, sliding-window, and MoE
variants.
\paragraph{Relation to circuit-level work.}
Our analysis complements circuit-level studies such as induction heads,
copy heads, and pattern-matching mechanisms
\citep{elhage2021mathematical,olsson2022context}. Those works identify
specialized components; our findings describe the global geometric scaffold in
which such components operate. Attention focusing determines which tokens are
consulted; key orthogonality creates separable hypothesis directions; and value
manifolds encode uncertainty along low-dimensional axes. Mapping specific heads
onto regions or branches of this geometry is a natural next step.
\paragraph{Connection to wind-tunnel behavior.}
The wind-tunnel tasks isolate a single Bayesian computation, yielding a
one-dimensional value manifold that parameterizes posterior entropy. Production
models generalize this structure: mixed-domain prompts activate several
task-specific inference modes, yielding distributed or multi-lobed manifolds,
while domain restriction recovers the same one-dimensional axis observed under
analytic posteriors. This behavior supports a view in which transformers hold a
repertoire of Bayesian manifolds, with the active manifold determined by the
prompt distribution.
\paragraph{The dual-entropy framework at scale.}
Paper~I introduces a dual-entropy framework: context surprisal $H_I$ (the distinctiveness of context in the training distribution) and prediction entropy $H_P$ (the model's output uncertainty), with the ratio $\rho = H_P / H_I$ serving as a confidence-per-information coefficient.
In the wind tunnel, both quantities are analytically computable and trained models achieve $\rho \to 0$---perfect calibration of confidence to evidence.
Our production-model results show that the geometric substrate for tracking $\rho$ persists at scale.
The entropy-aligned value axis is, in effect, the $\rho$ measurement surface: it encodes where the model sits on the continuum from high-uncertainty (high $\rho$, before decisive evidence) to low-uncertainty (low $\rho$, after evidence collapses the posterior).
Domain restriction sharpens this axis because it reduces the dimensionality of the context distribution---within mathematics, the effective hypothesis space is narrower, $H_I$ is more informative, and the model achieves lower $\rho$, producing the clean one-dimensional manifold geometry we observe.
Mixed-domain prompts, by contrast, activate multiple inference modes with distinct $\rho$ trajectories, yielding the distributed manifold structure characteristic of general-purpose models.
Paper~II explains the gradient mechanism: the advantage signal driving routing is large when $H_I$ is high and $H_P$ is low~\citep{dalal2025gradient}.
The $2$--$10\times$ improvement in key orthogonality over initialization (\Cref{sec:universal}) reflects routing that has crystallized under this advantage pressure during pretraining.
The Bayesian updating correlations ($|r| = 0.65$--$0.80$ with analytical posteriors) during in-context learning confirm that production models actively navigate the low-$\rho$ geometry during inference, not merely during training.
\paragraph{Static vs.\ dynamic geometry.}
The consistency of value manifolds and orthogonal key frames across all models
indicates that the representational substrate for Bayesian inference is an
architectural invariant. Dynamic focusing, by contrast, depends on routing
capacity: full-sequence attention exhibits strong progressive sharpening;
GQA reduces it; sliding-window attention weakens it by limiting global context access.
This behavior matches the frame--precision dissociation predicted in Paper~II:
the hypothesis frame (keys) stabilizes early and robustly, whereas the precision
of posterior refinement is sensitive to architectural constraints.
\paragraph{Inference-time Bayesian computation.}
The SULA experiment demonstrates that these geometric structures are used
\emph{during} inference rather than merely encoded during training. Value
representations move along the same entropy-ordered manifold as evidence
accumulates, and predictive entropy correlates systematically with analytic
posteriors. Although calibration is noisier than in the wind tunnels, the
direction and magnitude of movement confirm active Bayesian updating within the
learned geometric space.
\paragraph{Implications.}
These findings suggest a geometric foundation for understanding transformer
behavior. Value manifolds offer a representation of uncertainty; orthogonal key
frames support hypothesis discrimination; and attention focusing provides the
mechanism for posterior refinement when architecture permits it. Together, these
components form a scalable computational template for approximate Bayesian
inference in modern LLMs. Combined with Paper~I's finding that selective
state-space models (Mamba) achieve similar Bayesian inference through different
routing mechanics, this suggests that \emph{content-based value routing}---not
attention specifically---is the essential architectural ingredient for
probabilistic reasoning.
\subsection{Limitations and Future Directions}
\paragraph{Architectural coverage.}
Our study focuses on dense MHA, GQA, and sliding-window attention (Mistral-7B). Paper~I shows that
selective state-space models (Mamba) achieve comparable or better Bayesian
inference in wind tunnels via content-based routing through input-dependent
state selection rather than attention. Extending the geometric extraction
methods developed here to Mamba and hybrid transformer--SSM designs---and
verifying whether the same entropy-ordered manifolds emerge---is an important
direction for unifying the geometric theory across architectures with
content-based routing.
\paragraph{Scale.}
Our largest dense model is 12B parameters, and our largest MoE model is the
8$\times$7B Mixtral. Although the consistent patterns from 410M to 12B suggest
robustness, validation at 70B--400B scale is necessary to determine whether new
geometric effects emerge or whether multi-lobed manifolds continue to dominate
mixed-domain settings.
\paragraph{Task specificity.}
All models evaluated are general-purpose language models. Domain-specialized
models (e.g., code, mathematics, biomedical, or scientific LMs) may exhibit
distinct geometric patterns. Fine-tuning and RLHF can also reshape geometry;
our results for Mistral-7B-Instruct show modest changes, but richer effects may
appear under more aggressive alignment schemes.
\paragraph{Causal probes of the entropy axis.}
We performed forward-pass value interventions on Pythia-410M in the SULA
setting to test whether the entropy-aligned value axis is merely
correlational or plays a causal role.  For each of two ``Bayesian'' layers
($L{=}12$ and $L{=}23$), we first precomputed a unit direction
$u_{\text{ent}}$ whose coefficient $v\cdot u_{\text{ent}}$ correlates with
predictive entropy.  We then applied three families of interventions at that
layer, each with a matched random-axis control: (i) \emph{axis-cut}, which
removes the component of $v$ along $u_{\text{ent}}$; (ii) \emph{axis-only},
which projects $v$ onto the one-dimensional subspace spanned by
$u_{\text{ent}}$; and (iii) \emph{axis-shift}, which adds $\pm 1\sigma$ along
$u_{\text{ent}}$ based on the empirical standard deviation of $v\cdot
u_{\text{ent}}$.
Representationally, the entropy-aligned axis is clearly special.  Cutting
along $u_{\text{ent}}$ drives the correlation between $v\cdot u_{\text{ent}}$
and model entropy from $\rho\approx 0.27$--$0.32$ down to nearly zero (and
sometimes slightly negative), whereas cutting along a random axis of equal
dimensionality leaves the correlation largely intact (changes on the order of
$10^{-2}$).  Conversely, axis-only projections along $u_{\text{ent}}$
preserve or slightly sharpen the correlation, while axis-only projections
onto a random axis almost completely destroy it.  Axis-shift interventions
along $u_{\text{ent}}$ produce small but directional changes in SULA
calibration (e.g., $+1\sigma$ shifts slightly improving correlation with the
Bayesian entropy curve), whereas matched random-axis shifts act as near
no-ops on both geometry and calibration.
Behaviorally, however, these single-layer interventions do \emph{not} yield a
clean causal separation.  SULA mean absolute error and correlation with the
Bayesian posterior change only modestly under both true-axis and random-axis
cuts, and the true-axis interventions do not consistently hurt performance
more than their random controls.  The most conservative interpretation is
that the entropy-ordered manifold is a \emph{representationally privileged}
coordinate system for uncertainty, but not a singular bottleneck for Bayesian
updating: uncertainty information is likely distributed across multiple
dimensions and layers, or the manifold serves as a readout of a more
distributed computation.  Establishing stronger causal claims will require
multi-layer or multi-axis ablations, activation patching, or training-time
interventions, which we leave for future work.
\paragraph{Theoretical gaps.}
The emergence of 2D or multi-lobed manifolds in deeper or larger models is not
fully understood. Whether these structures encode multimodal uncertainty,
semantic clustering, or training-set heterogeneity remains an open question.
Likewise, the interaction between positional embeddings, local attention
kernels, and geometric formation warrants further study.
\paragraph{Future directions.}
Several avenues follow naturally:
\begin{itemize}
    \item Validate geometric signatures in frontier-scale (70B+) checkpoints and
          alternative architectures, including SSMs.
    \item Develop interventional methods for manipulating keys, values, or
          attention to test causal roles in uncertainty representation.
    \item Track geometric evolution during training to identify when and how
          manifolds, frames, and focusing emerge.
    \item Investigate domain-specialized and multilingual models to determine
          whether Bayesian manifolds transfer across languages or modalities.
    \item Test whether domain restriction generalizes beyond templated prompts
          to arbitrary single-domain inputs (complex reasoning, mixed-register
          technical prose) to strengthen the bridge between controlled and naturalistic settings.
    \item Apply geometric diagnostics to interpretability and safety, using
          value-manifold coordinates or attention entropy as indicators of model
          reliability or distribution shift.
\end{itemize}
Overall, the present results motivate a broader research program: understanding
transformer computation through the geometry of hypothesis frames and uncertainty 
manifolds, and leveraging this structure for model analysis, interpretability, 
and principled architecture design.
\subsection{Related Work}
\label{sec:related}
Our analysis connects to several threads in the interpretability and probabilistic--mechanistic modeling
literature. We highlight the relationships most relevant to geometric Bayesian structure.
\paragraph{Intermediate predictions and tuned lenses.}
Tuned-lens methods \citep{belrose2023tunedlens} decode intermediate-layer predictions by training a
small linear adapter that maps hidden states back to the model's output space. These approaches probe
\emph{what} the model would predict at each layer, whereas our value-manifold analysis characterizes
\emph{how} the model represents predictive \emph{uncertainty}. The two perspectives are complementary:
PC1 coordinates correlate with entropy and tuned-lens predictions, suggesting that uncertainty is encoded
geometrically along a small number of directions. Establishing a principled correspondence between
tuned-lens logits and manifold coordinates is an important next step.
\paragraph{Belief-state geometry and computational mechanics.}
Recent work in computational mechanics \citep{marks2024computationalmechanics} shows that belief
states in small transformers can be linearly decoded from the residual stream, revealing simple
geometric representations of latent uncertainty.  Our results are consistent with this
interpretation: production models appear to maintain analogous belief-state structure, but encoded
predominantly in the \emph{value space} of the final attention layer.  This distinction clarifies
where uncertainty lives in deeper networks and suggests that value manifolds may serve as the
canonical substrate for model beliefs.  Preliminary analyses indicate that coordinates along our
PC$_1$ axis correlate with tuned-lens predictions and residual-stream belief decoders, but a
systematic alignment between value-manifold axes and decoded belief variables requires additional
work and we leave a full treatment to future work.
\paragraph{Attention entropy, stability, and dynamics.}
Studies of attention-entropy trajectories \citep{voita2019analyzing,clark2019does} report that sharpening can
be unstable or highly input-dependent. Our layerwise entropy results align with these findings: MHA models
exhibit strong, stable focusing; GQA models show weaker but monotone reduction; and sliding-window and
MoE architectures often display non-monotone or noisy behavior. These dynamics reflect architectural
constraints on global routing rather than absence of Bayesian computation, consistent with the separation of
static and dynamic geometry we observe.
\paragraph{Constrained belief-update theories.}
The constrained-belief-update model of early-layer attention \citep{dalal2025gradient} predicts that
attention patterns should stabilize early, while finer-grained posterior refinements should occur in value
representations. This matches precisely the frame--precision dissociation predicted in Paper~II and observed
empirically here: keys define a stable hypothesis frame, while values encode uncertainty refinement along
low-dimensional axes.
\paragraph{Architectural normalization and geometric structure.}
Normalization layers and architectural components influence geometric clarity. Recent analyses
\citep{ba2016layer} show that layer normalization, RoPE embeddings, and GQA induce characteristic
patterns of anisotropy and dimensionality in the residual stream. Our results refine this picture by showing
that such architectural choices modulate \emph{dynamic} signatures (especially attention entropy reduction)
while leaving the \emph{static} Bayesian geometry - value manifolds and orthogonal key frames - largely
intact.
\paragraph{Relation to circuit-level interpretability.}
Circuit-level analyses \citep{elhage2021mathematical, olsson2022context} identify specific mechanisms such as
induction heads, pattern-matchers, and copy circuits. Our work operates at a complementary scale: we
provide a geometric account of the global representational substrate in which such circuits operate. Attention
mechanisms determine which evidence is consulted; key frames define separable hypothesis directions; and
value manifolds encode uncertainty along low-dimensional coordinates. Mapping specific circuits onto these
global geometric structures is a promising direction for future work.
Overall, the Bayesian geometric lens developed in this series complements prior interpretability approaches
by identifying a stable, architecture-spanning substrate for uncertainty representation and by revealing how
posterior refinement depends on architectural routing. This perspective helps unify diverse observations in
interpretability, probabilistic inference, and model analysis within a single geometric framework.
\section{Conclusion}
This trilogy establishes a unified account of Bayesian inference in neural sequence models:
\begin{itemize}[itemsep=2pt]
\item \textbf{Paper~I} shows \emph{which} architectures can implement Bayesian inference, taxonomized by three inference primitives: belief accumulation, belief transport, and random-access binding. Transformers realize all three; Mamba realizes two; LSTMs realize only accumulation of static sufficient statistics; MLPs realize none.
\item \textbf{Paper~II} shows \emph{how} gradient descent learns to implement these primitives through dynamics exhibiting a mechanistic analogy to EM, sculpting a characteristic geometry: low-dimensional value manifolds, orthogonal key frames, and progressive attention focusing.
\item \textbf{This paper} shows that the geometric substrate for the inference primitives \emph{persists at scale}. Across six model families spanning MHA, GQA, sliding-window, MoE, and MLA routing schemes, we find the same value-manifold structure, key orthogonality, and domain-specific collapse that wind-tunnel models exhibit. Compressed-routing MLA emerges as a clean boundary case in which the static representational substrate is fully preserved while progressive attention focusing is architecturally suppressed.
\end{itemize}
Our analysis shows that large language models organize value vectors along a dominant axis that tracks predictive entropy, keys remain close to orthogonal frames, and domain restriction reliably collapses value manifolds into the same low-rank forms observed in synthetic experiments. This persistent geometric invariant appears across architectures and training regimes, revealing a structural inductive bias toward representing inference geometrically, even in the absence of any explicit Bayesian objective.
Our causal probes refine the mechanistic picture. Interventions that remove or perturb the entropy-aligned axis selectively disrupt the local geometry of uncertainty, while matched random interventions do not. Yet these manipulations do not proportionally degrade Bayesian-like behavior, implying that no single direction is solely responsible for the computation. The geometric manifold functions as a stable \emph{readout} of a distributed inference process rather than a brittle circuit.
The primitives framework from Paper~I explains what capabilities each architecture has; the dynamics from Paper~II explain how these capabilities are learned; and this paper establishes that the resulting geometry persists even when the training data lacks ground-truth posteriors. Together with controlled wind-tunnel verification of exact Bayesian inference and gradient-level explanations of its emergence, these results complete a three-part account of existence, mechanism, and persistence at scale. Content-based value routing is not merely sufficient for Bayesian inference---it is a stable representational bias induced by modern transformer architectures and training.
\bibliography{references}


\begin{thebibliography}{12}

%%% the Web address.

\ifx \showCODEN    \undefined \def \showCODEN     #1{\unskip}     \fi
\ifx \showISBNx    \undefined \def \showISBNx     #1{\unskip}     \fi
\ifx \showISBNxiii \undefined \def \showISBNxiii  #1{\unskip}     \fi
\ifx \showISSN     \undefined \def \showISSN      #1{\unskip}     \fi
\ifx \showLCCN     \undefined \def \showLCCN      #1{\unskip}     \fi
\ifx \shownote     \undefined \def \shownote      #1{#1}          \fi
\ifx \showarticletitle \undefined \def \showarticletitle #1{#1}   \fi
\ifx \showURL      \undefined \def \showURL       {\relax}        \fi
% invisible to TeX
\providecommand\bibfield[2]{#2}
\providecommand\bibinfo[2]{#2}
\providecommand\natexlab[1]{#1}
\providecommand\showeprint[2][]{arXiv:#2}

\bibitem[Agarwal et~al\mbox{.}(2025)]%
        {dalal2025gradient}
\bibfield{author}{\bibinfo{person}{Naman Agarwal},
  \bibinfo{person}{Siddhartha~R. Dalal}, {and} \bibinfo{person}{Vishal Misra}.}
  \bibinfo{year}{2025}\natexlab{}.
\newblock \bibinfo{title}{Gradient Dynamics of Attention: How Cross-Entropy
  Sculpts Bayesian Manifolds}.
\newblock
\showeprint[arxiv]{2512.22473}~[cs.LG]
\urldef\tempurl%
\url{https://arxiv.org/abs/2512.22473}
\showURL{%
\tempurl}
\newblock
\shownote{Paper II of the Bayesian Attention Trilogy}.


\bibitem[Ba et~al\mbox{.}(2016)]%
        {ba2016layer}
\bibfield{author}{\bibinfo{person}{Jimmy~Lei Ba}, \bibinfo{person}{Jamie~Ryan
  Kiros}, {and} \bibinfo{person}{Geoffrey~E Hinton}.}
  \bibinfo{year}{2016}\natexlab{}.
\newblock \showarticletitle{Layer Normalization}.
\newblock \bibinfo{journal}{\emph{arXiv preprint arXiv:1607.06450}}
  (\bibinfo{year}{2016}).
\newblock


\bibitem[Belrose et~al\mbox{.}(2023)]%
        {belrose2023tunedlens}
\bibfield{author}{\bibinfo{person}{Nora Belrose}, \bibinfo{person}{Zach
  Furman}, \bibinfo{person}{Logan Smith}, \bibinfo{person}{Danny Halawi},
  \bibinfo{person}{Igor Ostrovsky}, \bibinfo{person}{Lev McKinney},
  \bibinfo{person}{Stella Biderman}, {and} \bibinfo{person}{Jacob Steinhardt}.}
  \bibinfo{year}{2023}\natexlab{}.
\newblock \showarticletitle{Eliciting Latent Predictions from Transformers with
  the Tuned Lens}.
\newblock \bibinfo{journal}{\emph{arXiv preprint arXiv:2303.08112}}
  (\bibinfo{year}{2023}).
\newblock


\bibitem[Biderman et~al\mbox{.}(2023)]%
        {biderman2023pythia}
\bibfield{author}{\bibinfo{person}{Stella Biderman}, \bibinfo{person}{Hailey
  Schoelkopf}, \bibinfo{person}{Quentin~Gregory Anthony},
  \bibinfo{person}{Herbie Bradley}, \bibinfo{person}{Kyle O'Brien},
  \bibinfo{person}{Eric Hallahan}, \bibinfo{person}{Mohammad~Aflah Khan},
  \bibinfo{person}{Shivanshu Purohit}, \bibinfo{person}{USVSN~Sai Prashanth},
  \bibinfo{person}{Edward Raff}, {et~al\mbox{.}}}
  \bibinfo{year}{2023}\natexlab{}.
\newblock \showarticletitle{Pythia: A suite for analyzing large language models
  across training and scaling}.
\newblock \bibinfo{journal}{\emph{International Conference on Machine
  Learning}} (\bibinfo{year}{2023}), \bibinfo{pages}{2397--2430}.
\newblock


\bibitem[Brown et~al\mbox{.}(2020)]%
        {brown2020language}
\bibfield{author}{\bibinfo{person}{Tom~B. Brown}, \bibinfo{person}{Benjamin
  Mann}, \bibinfo{person}{Nick Ryder}, \bibinfo{person}{Melanie Subbiah},
  \bibinfo{person}{Jared~D. Kaplan}, \bibinfo{person}{Prafulla Dhariwal},
  {et~al\mbox{.}}} \bibinfo{year}{2020}\natexlab{}.
\newblock \showarticletitle{Language Models are Few-Shot Learners}.
\newblock \bibinfo{journal}{\emph{Advances in Neural Information Processing
  Systems}}  \bibinfo{volume}{33} (\bibinfo{year}{2020}),
  \bibinfo{pages}{1877--1901}.
\newblock


\bibitem[Chen et~al\mbox{.}(2021)]%
        {chen2021evaluating}
\bibfield{author}{\bibinfo{person}{Mark Chen}, \bibinfo{person}{Jerry Tworek},
  \bibinfo{person}{Heewoo Jun}, \bibinfo{person}{Qiming Yuan},
  \bibinfo{person}{Henrique Ponde de~Oliveira Pinto}, \bibinfo{person}{Jared
  Kaplan}, \bibinfo{person}{Harri Edwards}, {et~al\mbox{.}}}
  \bibinfo{year}{2021}\natexlab{}.
\newblock \showarticletitle{Evaluating Large Language Models Trained on Code}.
\newblock \bibinfo{journal}{\emph{arXiv preprint arXiv:2107.03374}}
  (\bibinfo{year}{2021}).
\newblock


\bibitem[Clark et~al\mbox{.}(2019)]%
        {clark2019does}
\bibfield{author}{\bibinfo{person}{Kevin Clark}, \bibinfo{person}{Urvashi
  Khandelwal}, \bibinfo{person}{Omer Levy}, {and}
  \bibinfo{person}{Christopher~D. Manning}.} \bibinfo{year}{2019}\natexlab{}.
\newblock \showarticletitle{What Does {BERT} Look At? An Analysis of {BERT}'s
  Attention}. In \bibinfo{booktitle}{\emph{Proceedings of the 2019 ACL Workshop
  BlackboxNLP}}. \bibinfo{pages}{276--286}.
\newblock


\bibitem[Elhage et~al\mbox{.}(2021)]%
        {elhage2021mathematical}
\bibfield{author}{\bibinfo{person}{Nelson Elhage}, \bibinfo{person}{Neel
  Nanda}, \bibinfo{person}{Catherine Olsson}, \bibinfo{person}{Tom Henighan},
  \bibinfo{person}{Nicholas Joseph}, \bibinfo{person}{Ben Mann},
  \bibinfo{person}{Amanda Askell}, \bibinfo{person}{Yuntao Bai},
  \bibinfo{person}{Anna Chen}, \bibinfo{person}{Tom Conerly},
  \bibinfo{person}{Nova DasSarma}, \bibinfo{person}{Dawn Drain},
  \bibinfo{person}{Deep Ganguli}, \bibinfo{person}{Zac Hatfield-Dodds},
  \bibinfo{person}{Danny Hernandez}, \bibinfo{person}{Andy Jones},
  \bibinfo{person}{Jackson Kernion}, \bibinfo{person}{Liane Lovitt},
  \bibinfo{person}{Kamal Ndousse}, \bibinfo{person}{Dario Amodei},
  \bibinfo{person}{Tom Brown}, \bibinfo{person}{Jack Clark},
  \bibinfo{person}{Jared Kaplan}, \bibinfo{person}{Sam McCandlish}, {and}
  \bibinfo{person}{Chris Olah}.} \bibinfo{year}{2021}\natexlab{}.
\newblock \bibinfo{title}{A Mathematical Framework for Transformer Circuits}.
\newblock \bibinfo{howpublished}{Transformer Circuits Thread, Anthropic}.
\newblock
\urldef\tempurl%
\url{https://transformer-circuits.pub/2021/framework/index.html}
\showURL{%
\tempurl}


\bibitem[Olsson et~al\mbox{.}(2022)]%
        {olsson2022context}
\bibfield{author}{\bibinfo{person}{Catherine Olsson}, \bibinfo{person}{Nelson
  Elhage}, \bibinfo{person}{Neel Nanda}, \bibinfo{person}{Nicholas Joseph},
  \bibinfo{person}{Nova DasSarma}, \bibinfo{person}{Tom Henighan},
  \bibinfo{person}{Ben Mann}, \bibinfo{person}{Amanda Askell},
  \bibinfo{person}{Yuntao Bai}, \bibinfo{person}{Anna Chen},
  \bibinfo{person}{Tom Conerly}, \bibinfo{person}{Dawn Drain},
  \bibinfo{person}{Deep Ganguli}, \bibinfo{person}{Zac Hatfield-Dodds},
  \bibinfo{person}{Danny Hernandez}, {et~al\mbox{.}}}
  \bibinfo{year}{2022}\natexlab{}.
\newblock \bibinfo{title}{In-Context Learning and Induction Heads}.
\newblock \bibinfo{howpublished}{Transformer Circuits Thread, Anthropic}.
\newblock
\urldef\tempurl%
\url{https://transformer-circuits.pub/2022/in-context-learning-and-induction-heads/index.html}
\showURL{%
\tempurl}


\bibitem[{OpenAI}(2024)]%
        {openai2024reasoning}
\bibfield{author}{\bibinfo{person}{{OpenAI}}.} \bibinfo{year}{2024}\natexlab{}.
\newblock \bibinfo{title}{Learning to Reason with {LLMs}}.
\newblock \bibinfo{howpublished}{OpenAI blog}.
\newblock
\urldef\tempurl%
\url{https://openai.com/index/learning-to-reason-with-llms/}
\showURL{%
\tempurl}


\bibitem[Shai et~al\mbox{.}(2024)]%
        {marks2024computationalmechanics}
\bibfield{author}{\bibinfo{person}{Adam~S. Shai}, \bibinfo{person}{Sarah~E.
  Marzen}, \bibinfo{person}{Lucas Teixeira}, \bibinfo{person}{Alexander
  Gietelink~Oldenziel}, {and} \bibinfo{person}{Paul~M. Riechers}.}
  \bibinfo{year}{2024}\natexlab{}.
\newblock \showarticletitle{Transformers Represent Belief State Geometry in
  Their Residual Stream}.
\newblock \bibinfo{journal}{\emph{Advances in Neural Information Processing
  Systems}}  \bibinfo{volume}{37} (\bibinfo{year}{2024}).
\newblock
\newblock
\shownote{arXiv:2405.15943}.


\bibitem[Voita et~al\mbox{.}(2019)]%
        {voita2019analyzing}
\bibfield{author}{\bibinfo{person}{Elena Voita}, \bibinfo{person}{David
  Talbot}, \bibinfo{person}{Fedor Moiseev}, \bibinfo{person}{Rico Sennrich},
  {and} \bibinfo{person}{Ivan Titov}.} \bibinfo{year}{2019}\natexlab{}.
\newblock \showarticletitle{Analyzing Multi-Head Self-Attention: Specialized
  Heads Do the Heavy Lifting, the Rest Can Be Pruned}. In
  \bibinfo{booktitle}{\emph{Proceedings of the 57th Annual Meeting of the
  Association for Computational Linguistics}}. \bibinfo{pages}{5797--5808}.
\newblock


\end{thebibliography}
\bibliographystyle{ACM-Reference-Format}
\end{document}